\documentclass{article}
\usepackage{ijcai24}

\usepackage{times}
\usepackage{soul}
\usepackage{url}
\usepackage[hidelinks,allcolors=blue]{hyperref}
\usepackage[utf8]{inputenc}
\usepackage[small]{caption}
\usepackage{graphicx}
\usepackage{amsmath}
\usepackage{amsthm}
\usepackage{booktabs}
\usepackage{algorithm}
\usepackage{algorithmic}
\usepackage[switch]{lineno}
\usepackage{threeparttable}
\usepackage{multirow}
\usepackage{xcolor}
\usepackage{amssymb}
\usepackage{subfig}

\newtheorem{definition}{Definition}

\title{MINI-LLM: Memory-Efficient Structured Pruning for Large Language Models}

\author{
Hongrong Cheng$^1$
\and
Miao Zhang$^2$\footnote{Corresponding author.}\and
Javen Qinfeng Shi$^{1}$
\affiliations
$^1$University of Adelaide, Adelaide, Australia\\
$^2$Harbin Institute of Technology, Shenzhen, China\\
\emails
\{hongrong.cheng, javen.shi\}@adelaide.edu.au,
zhangmiao@hit.edu.cn
}

\begin{document}

\maketitle

\begin{abstract}
As Large Language Models (LLMs) grow dramatically in size, there is an increasing trend in compressing and speeding up these models. Previous studies have highlighted the usefulness of gradients for importance scoring in neural network compressing, especially in pruning medium-size networks. However, the substantial memory requirements involved in calculating gradients with backpropagation impede the utilization of gradients in guiding LLM pruning. As a result, most pruning strategies for LLMs rely on gradient-free criteria, such as weight magnitudes or a mix of magnitudes and activations. In this paper, we devise a hybrid pruning criterion, which appropriately integrates magnitude, activation, and gradient to capitalize on feature map sensitivity for pruning LLMs. To overcome memory requirement barriers, we estimate gradients using only forward passes. Based on this, we propose a Memory-effIcieNt structured prunIng procedure for LLMs (MINI-LLM) to remove no-critical channels and multi-attention heads. Experimental results demonstrate the superior performance of MINI-LLM over existing gradient-free methods on three LLMs: LLaMA, BLOOM, and OPT across various downstream tasks (classification, multiple-choice, and generation), while MINI-LLM maintains a GPU memory footprint akin to gradient-free methods. 
\end{abstract}

\section{Introduction}
The advent of pre-trained Large Language Models (LLMs), such as GPT-4~\cite{open2023gpt4} and LLaMA~\cite{touvron2023llama}, has made remarkable processes across various complex Natural Language Processing (NLP) tasks, such as natural language generation~\cite{wu2020debiased}, question answering~\cite{brown2020language}, and recommendation system~\cite{wu2023survey}. However, this remarkable capability usually entails a large model size, resulting in significant computational costs in terms of storage, memory, and computation time, which presents considerable difficulties during the training and deployment phases. To this end, there has been considerable interest in compressing LLMs~\cite{ma2023llmpruner,dettmers2023qlora,frantar2023sparsegpt,xiao2023smoothquant,li2020group} to make them more practical for various tasks. 
Neural network pruning~\cite{ma2023llmpruner,frantar2023sparsegpt,sun2024simple,xia2024sheared}, as one of the indispensable approaches for compressing and accelerating neural networks, has recently found its way into LLMs. 

\begin{figure}[t]
\centering
\begin{minipage}{6.9cm}
  \includegraphics[width=6.9cm,height=4cm]{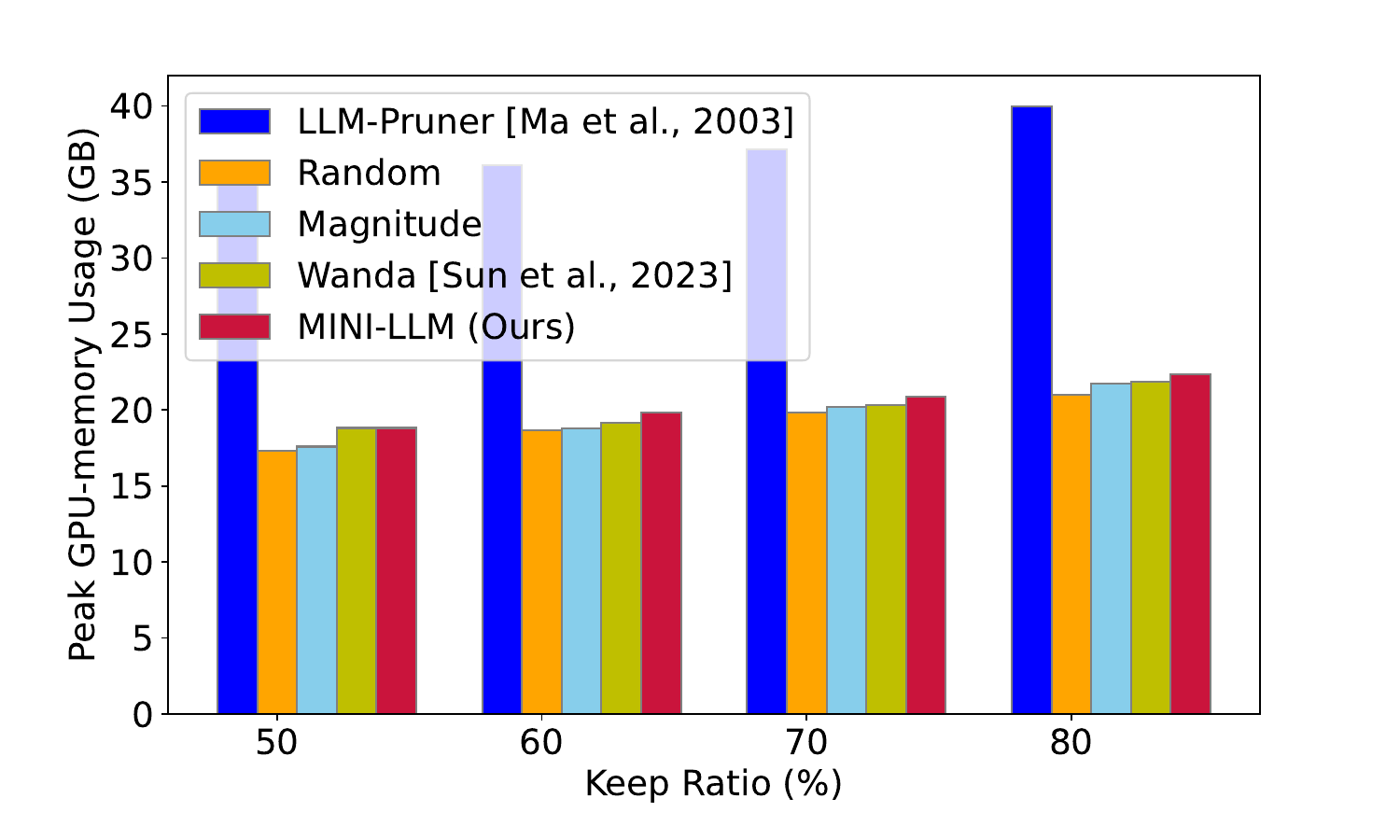}
\end{minipage}
\caption{The peak GPU-memory Usage for pruning LLaMA-7B. The backpropagation gradient-based pruning method, LLM-Pruner, consumes about twice the GPU resources compared to gradient-free methods and our method MINI-LLM during pruning LLaMA-7B.}
\label{Fig:peak-GPU-pruning-llama-7b}
\vspace{-0.3cm}
\end{figure}

In the traditional pruning methods~\cite{molchanov2017pruning,lee2019snip,sanh2020movement,liu2021group,fu2022depthshrinker} for compressing small or medium-size models, gradients of loss functions w.r.t. weights, masks, or feature maps have demonstrated more reliable performance than gradient-free methods (e.g., magnitude-based methods) in discriminating important weights/channels. For example, \cite{lee2019snip} exploits the first-order Taylor expansion to identify the connection sensitivity caused by setting some weights to zero, which outperforms magnitude-based methods and obtains extremely sparse networks with similar accuracy as the reference networks. However, due to the huge number of parameters in LLMs, computing gradients with backpropagation requires a prohibitive amount of memory. LLM-Pruner~\cite{ma2023llmpruner}, which uses gradients calculated via backpropagation for structured pruning, consumes about twice the GPU resources compared to magnitude-based methods during pruning LLaMA-7B, as illustrated in Figure~\ref{Fig:peak-GPU-pruning-llama-7b}\footnote{We obtained the data on an NVIDIA A100 (40GB)}. Even though the recovery stage can be executed on a single RTX 4090 (24GB) with the help of Low-Rank Adaption (LoRA)~\cite{hu2022lora}, the GPU consumption during the pruning stage has become the bottleneck for GPU resource usage in the entire pruning framework.  

To avoid incurring untenable memory costs for computing gradients on LLMs, some pruning methods~\cite{frantar2023sparsegpt,sun2024simple} constructed gradient-free criteria. For example, SparseGPT~\cite{frantar2023sparsegpt} employed the combination of weight magnitude and the inverse Hessian matrix which is formed from the product of given input features to score weights for unstructured pruning. However, the computation of the inverse Hessian matrix is resource-intensive, and the utility of gradient information remains under-exploited. Some pruning methods, such as Sheared LLaMA [\cite{xia2024sheared}], combine pruning with pre-training, which requires substantial GPU resources. For example, Sheared LLaMA requires 8 A100 (80GB) GPUs for pruning and 16 for pre-training. Since this paper focuses on memory-efficient pruning methods and follows the mainstream framework that starts with pruning and then fine-tuning, without pre-training, although outstanding in performance, these methods are not within the scope of our discussion. 

In this paper, we propose the Memory-effIcieNt structured prunIng procedure for LLMs (MINI-LLM), which scores weights using estimated gradients with only forward passes. We make this approach tractable by contributing multiple techniques. 

Our main contributions are as follows:
\begin{itemize}
\item We design a novel pruning criterion called Feature Map Sensitivity (FMS) score, integrating weight magnitude, activation, and gradient. This criterion optimally utilizes the pivotal information from the three critical aspects, which facilitates a more nuanced assessment of feature map sensitivity and provides effective scoring in LLMs.

\item We propose a structured pruning framework for LLMs called MINI-LLM which utilizes estimated gradients with only forward passes by using comparable GPU memory usage to gradient-free methods, significantly improving GPU memory efficiency over traditional backpropagation gradients.
 
\item The experiments on three types of LLMs: LLaMA, BLOOM, and OPT over different downstream tasks (classification, multiple-choices, and generation) demonstrate that the novel pruning criterion FMS can effectively boost the performance of gradient-based methods. Additionally, our proposed gradient-based structured pruning method MINI-LLM steadily exceeds gradient-free pruning methods in performance and rivals or surpasses backpropagation gradient-based method at times, while using similar GPU memory as gradient-free methods.
 
\end{itemize}

\section{Related Work}
\textbf{Structured/Unstructured/Semi-structured LLM pruning.}
The pruning methods for LLMs can still be generally categorized as unstructured (\cite{sun2024simple,frantar2022optimal}), semi-structured (\cite{frantar2023sparsegpt}), and structured (\cite{ma2023llmpruner,wang2020structured}) pruning methods, similar to the categorization for pruning small and mid-size neural networks. Unstructured pruning achieves substantial sparsity by directly setting weights or their masks to zero while maintaining a comparable performance compared to the vanilla models. However, the irregular sparsity results in no compression in the model size, and actual acceleration necessitates the support of specialized software/hardware. In contrast, structured pruning discards the whole grouped parameters (such as channels and attention heads), leading to physically reduced model size and enabling inference acceleration without any special requirements of software/hardware (\cite{zhou2022transpim,frantar2023sparsegpt}). Semi-structure pruning, such as 2:4 or 4:8 patterns in \cite{frantar2023sparsegpt}, provides a balance between performance and hardware speedup. In this paper, we focus on structured pruning for LLMs.

\vspace{0.2cm}
\noindent \textbf{Pruning criteria for LLMs.}
Neural network pruning methods search for an optimal subnetwork by removing unimportant weights. As one of the most popular criterion factors, gradients have already been demonstrated effective in constructing scoring functions for pruning small or medium-size networks~\cite{liu2021group,fu2022depthshrinker,wang2020picking,yu2022combinatorial,molchanov2019importance,kwon2022fast}. However, calculating gradients using backpropagation is highly resource-intensive for GPU memory, making it challenging to implement for LLMs, where meeting such high memory demands is difficult. For example, LLM-Pruner~\cite{ma2023llmpruner} employs gradients calculated through backpropagation for LLM pruning, but the GPU memory required for pruning exceeds that of fine-tuning. To this end, there are some gradient-free pruning methods (\cite{frantar2023sparsegpt,sun2024simple,nova2023gradient,kurtic2023ziplm,li2022parameter}). Most of them are centered on post-training (retraining-free) approaches that involve pruning while concurrently compensating for performance. For instance, 
Wanda~\cite{sun2024simple} multiplies weight magnitude and the corresponding activation to implement unstructured post-training pruning for LLMs. Even though these gradient-free methods are GPU memory efficient, the utility of gradient remains under-exploited. This paper actively seeks an effective estimation method to overcome memory requirement barriers related to computing gradients with backpropagation. 

\vspace{0.2cm}
\noindent \textbf{Zeroth-Order optimization.} 
Zeroth-Order (ZO) optimization can fall in the general class of weight perturbation methods. An early method referred to as the Finite Difference Stochastic Approximation (FDSA) \cite{kiefer1952stochastic} estimated the gradient by using $2d$ function measurements, two for each of the $d$ partial derivatives. One of the key disadvantages of FDSA is that a large $d$ value would result in serious computational challenges. A more efficient gradient estimation method is Simultaneous Perturbation Stochastic Approximation (SPSA)~\cite{spall1992multivariate,li2022simultaneous} which approximates gradients using only two forward passes. In previous work, ZO gradient estimation was used for solving optimization-related problems, such as for model training. For example, Malladi et al.~\cite{malladi2023finetuning} propose a Memory-efficient ZO-SGD (MeZO) to adapt SPSA to fine-tuning LLMs in a memory-efficient way. In this paper, for the first time, we apply ZO gradient estimation for pruning LLMs.

\section{Method}
In this section, we propose a Memory-effIcieNt structured prunIng procedure for LLMs termed MINI-LLM. We start by describing a new pruning criterion that evaluates feature map saliency from three critical factors: gradient, weight magnitude, and activation. To evaluate gradients in a memory-efficient way, we exploit ZO gradients to approximate the backpropagation based gradients. Finally, to recover performance, we utilize LoRA~\cite{hu2022lora} to fine-tune the pruned model, which has high training throughput, but low GPU memory requirement. 
    
\subsection{Pruning Criterion in MINI-LLM}
\label{sec:our-pruning-criterion}
\textbf{Problem Definition.} The pruning problem for LLMs starts from a pre-trained dense model $W_{0}\in \mathbb{R}^{d}$ and aims to find a sparse version of $W_{0}$, where many channels and attention heads are discarded. The remaining weights $\hat{W}_{0}$ may be updated accordingly to preserve the performance. Consider a labeled dataset $\mathcal{D}=\{(x_{i},y_{i})\}_{i=1}^{N}$, where $N$ is the number of samples, and a desired prune ratio $p$ (i.e., the percentage of removed weights), our goal is to remove the weights that has the least impact on the model's prediction. Therefore, neural network pruning can be formulated as the following constrained optimization problem:
\begin{equation}
\begin{aligned}
&\mathop{\textrm{min}}\limits_{\hat{W}_{0}} \ \mathcal{L}(\hat{W}_{0};\mathcal{D})=\mathop{\textrm{min}}\limits_{\hat{W}_{0}}\frac{1}{N}\sum_{i=1}^{N}\ell(\hat{W}_{0};(x_{i},y_{i})), \\
&\text{s.t.} \quad \lvert |\hat{W}_{0} \rvert |_{0} \leq \lvert| W_{0} \rvert |_{0} \times (1 - p), \\
\end{aligned}
\end{equation}
where $\ell(\cdot)$ can be the standard loss function (e.g., cross-entropy loss) and $\lvert | \cdot \rvert|_{0}$ is the standard $L_{0}$ norm. 

\vspace{0.2cm}
\noindent \textbf{Gradient-based Pruning.} To evaluate the significance of a specific weight $W_{l}^{k}$, one common way is using its sensitive effect on the loss function, where $k$ denotes the $k$-th weight in the $l$-th layer. Specifically, one can compare the difference in the loss function when $W_{l}^{k}$ is included versus when it is excluded from the model (i.e., LLaMA-7B). Thus, the loss change can be formulated as~\cite{lecun1989optimal}: 
\begin{equation}
\begin{aligned}
\Delta \mathcal{L} &= \mathcal{L}_{W_{l}^{k}}(\mathcal{D}) - \mathcal{L}_{W_{l}^{k}=0}(\mathcal{D}) \\
&\approx \Delta W^{T}\frac{\partial \mathcal{L}(\mathcal{D})}{\partial W_{l}^{k}} - \frac{1}{2}\Delta W^{T} H \Delta W \\
&= W_{l}^{k}\frac{\partial \mathcal{L}(\mathcal{D})}{\partial W_{l}^{k}} - \frac{1}{2}W_{l}^{k}H W_{l}^{k}, \\
\end{aligned}
\end{equation}
where $\Delta W=W_{l}^{k}$ and the Hessian matrix $H=\nabla^{2}_{W_{l}^{k}}\mathcal{L}(W_{0})$. Unlike previous work~\cite{liu2021group,kurtic2022optimal}, the pre-trained datasets of a large language model are inconsistent with the datasets of the downstream tasks, hence $\frac{\partial \mathcal{L}(\mathcal{D})}{\partial W_{l}^{k}} \not \approx 0$. This characteristic is advantageous for assessing the importance of weights through the gradient term in the context of LLMs, as calculating the Hessian matrix in the second term is impractical on LLMs with $\mathcal{O}(N^{2})$ complexity. Therefore, $\Delta \mathcal{L}$ can be approximated as:
\begin{equation}
\Delta \mathcal{L} = \mathcal{L}_{W_{l}^{k}}(\mathcal{D}) - \mathcal{L}_{W_{l}^{k}=0}(\mathcal{D}) \approx W_{l}^{k}\frac{\partial \mathcal{L}(\mathcal{D})}{\partial W_{l}^{k}}.
\label{eq:sensitivity-score-taylor}
\end{equation}

\noindent \textbf{Activation-based Pruning.} As recently observed in LLMs larger than 6.7B (\cite{dettmers2022llmint8}), a small set of hidden state features emerges with significantly larger magnitudes (outliers) than the remainders and zeroing out these features causes a significant degradation of performance. The vanilla scoring function Eq.~\eqref{eq:sensitivity-score-taylor} does not highlight the unique characteristics of LLMs compared with smaller models. Given the $l$th layer's input activation $X_{l}$ (i.e., the output from the ($l-1$)th layer of the network) and the $l$th layer's weights $W_{l}$, the pruning problem is also commonly treated as finding the subset $\hat{W}_{l}$ of $W_{l}$ respecting a compression constraint $\mathcal{C}$, which most closely approximates the initial output as determined by the squared error metric. Assuming that the activation and weight matrices possess a suitable rectangular shape, the neural network pruning is defined as the following optimization problem \cite{frantar2023sparsegpt}:
\begin{equation}
\begin{aligned}
&\mathop{\textrm{min}}\limits_{\hat{W}_{l}}\ \lvert |\hat{W}_{l}X_{l}-W_{l}X_{l} \rvert|_{2}^{2} = \lvert | \Delta W_{l}X_{l}\rvert |_{2}^{2}, \\
&\text{s.t.} \quad \hat{W}_{l}\in \mathcal{C}.   \\ 
\end{aligned}
\end{equation}
To evaluate the significance of a specific weight $W_{l}^{k}$, one can compare the difference in the layer-wise output when $W_{l}^{k}$ is preserved versus when it is excluded from the model and write the formulation as:
\begin{equation}
\lvert | \Delta W_{l}X_{l}\rvert |_{2}=\left | W_{l}^{k}X_{l}^{k}\right |,
\label{eq:sensitivity-score-output}
\end{equation}
where $\Delta W_{l}=W_{l}^{k}$.

\vspace{0.2cm}
\noindent \textbf{Feature Map Sensitivity (FMS).} Eq.~\eqref{eq:sensitivity-score-output} only considers the output changes in a single layer and does not take into account the global loss change across the entire network. We notice that, with the weight gradients, the global loss change can be quantified with weight change as shown in Eq.~\eqref{eq:sensitivity-score-taylor}. To calculate the salience of each weight relative to the change in global loss and layer-wise output, we measure $\Delta W_{l}$ in Eq.~\eqref{eq:sensitivity-score-output} through its global sensitivity $\Delta \mathcal{L}$ from Eq.~\eqref{eq:sensitivity-score-taylor} and propose a heuristic hybrid sensitivity scoring function called Feature Map Sensitivity (\textbf{FMS}) as follows:
\begin{equation}
S(W_{l}^{k})_{ours}=\left | W_{l}^{k}\frac{\partial \mathcal{L}(\mathcal{D})}{\partial W_{l}^{k}}X_{l}^{k}\right |.
\label{eq:sensitivity-score-ours}
\end{equation}
Compared to Eq.~\eqref{eq:sensitivity-score-taylor} and Eq.~\eqref{eq:sensitivity-score-output}, our criterion Eq.~\eqref{eq:sensitivity-score-ours} integrates magnitude, activation, and gradient to optimally utilize the pivotal information from the three critical aspects, so as calculate the feature map sensitivity along with the loss changes.

\subsection{Pruning with Estimated Gradients}
Let $\mathcal{L}(W;\mathcal{B})$ denote the loss on a minibatch $\mathcal{B} \subset \mathcal{D}$. The following Definition~\ref{def-zo-gradient} describes a classical ZO gradient estimation based on SPSA (\cite{spall1992multivariate}).
\begin{definition}[ZO Gradient Estimation.]
    Given a model with parameters $W\in \mathbb{R}^{d}$ and a loss function $\mathcal L$, ZO gradient on a minibatch $\mathcal{B}$ is as
    \begin{equation}
    \resizebox{.9\linewidth}{!}{$
    \hat{\nabla}\mathcal{L}(W;\mathcal{B})=\frac{\mathcal{L}(W+\epsilon z; \mathcal{B})-\mathcal{L}(W-\epsilon z; \mathcal{B})}{2\epsilon z}\approx \nabla \mathcal{L}(W;\mathcal{B}),
    $}
    \label{eq:def-zo-gradient}
    \end{equation}
    \label{def-zo-gradient}
\end{definition}
\noindent where $\nabla \mathcal{L}(W;\mathcal{B})$ is the gradient with backpropagation, $z\in \mathbb{R}^{d}$ with $z \sim \mathcal{N}(0,I_{d})$ and $\epsilon$ is the perturbation scale. The $n$-ZO gradient estimate averages $\hat{\nabla}\mathcal{L}(W;\mathcal{B})$ over $n$ randomly sampled $z$. Malladi et al.~\shortcite{malladi2023finetuning} found that $n=1$ is the most efficient. Therefore, we choose $n=1$ as the default. For each weight $W_{l}^{k}$ in the model, the estimation of its gradient $\frac{\partial \mathcal{L}(\mathcal{D})}{\partial W_{l}^{k}}$ (defined as $\hat{g}_{l}^{k}$) is then
\begin{equation}
\hat{g}_{l}^{k}=\frac{\mathcal{L}(W+\epsilon z; \mathcal{B})-\mathcal{L}(W-\epsilon z; \mathcal{B})}{2\epsilon z_{l}^{k}},
\label{eq:zo-gradient-estimate}
\end{equation}
where $z_{l}^{k}\in z$ is the random corresponding to $W_{l}^{k}$. In this way, the practical pruning score used in our MINI-LLM is defined as:
\begin{equation}
\hat{S}(W_{l}^{k})_{ours}=\left | W_{l}^{k}\hat{g}_{l}^{k}X_{l}^{k}\right |.
\label{eq:sensitivity-score-ours-e}
\end{equation}  

\vspace{0.2cm}
\noindent \textbf{Dependency-aware structured LLM pruning.} To maintain structural integrity, it is crucial for structured pruning to identify groups of interdependent structures within LLMs. Following Ma et al.~\shortcite{ma2023llmpruner}, we prune heads for Multi-Head Attention (MHA) and channels for Feed-Forward Network (FFN), respectively. We arrange the interconnected weights into groups and determine the sensitivity of each group (a set of coupled structures) defined as $G=\{W_{i}\}_{i=1}^{M}$ by choosing the maximum sensitivity score of the structures in it, i.e., $\hat{S}({G})=\text{max}_{i=1}^{M}\sum_{k}\hat{S}(W_{i}^{k})$, where $M$ is the number of interdependent structures in the group. Our structured pruning approach MINI-LLM is outlined in Algorithm~\ref{alg:mini-llm}.

\begin{algorithm}[tb]
    \caption{The structured pruning algorithm MINI-LLM}
    \label{alg:mini-llm}
    \textbf{Input}: Dataset $\mathcal{D}$, pre-trained weights $W_{0}\in \mathbb{R}^{d}$, loss $\mathcal{L}:\mathbb{R}^{d} \rightarrow \mathbb{R}$, prune ratio $p$, perturbation scale $\epsilon$. \\
    \textbf{Output}: The pruned model
    \begin{algorithmic}[1] 
        \STATE Clear every weight's sensitivity score $\hat{S}(W_{l}^{k})=0$;
        \STATE Forward via Eq.~\eqref{eq:def-zo-gradient} and estimate each weight's  $\hat{g}_{l}^{k}$;
        \FOR{$l\in [1,...,L]$}
        \STATE Compute the input activation $X_{l}$ for $l$-th layer;
        \STATE Compute every weight's score $\hat{S}(W_{l}^{k})$ via Eq.~\eqref{eq:sensitivity-score-ours-e};        
        \ENDFOR
        \FOR{$l\in [1,...,L]$}
        \STATE Keep the important groups $\hat{S}(G_{l})$ ranked in top $1-p$; 
        \ENDFOR
        \STATE \textbf{return} the pruned model.
    \end{algorithmic}
\end{algorithm}

\subsection{Recovery with Low-rank Approximation}
After pruning, we need a recovery stage to regain the performance. Due to the huge number of parameters, full fine-tuning becomes less feasible. LoRA~\cite{hu2022lora}, as one of the most popular Parameter-Efficient Fine-Tuning (PEFT) methods~\cite{he2023sensitivity,li2021prefixtuning,jia2022visual,chavan2023oneforall,lester2021power}, has demonstrated strong capability for performance recovery, while significantly reducing GPU memory usage \cite{dettmers2023qlora}.

To facilitate this, we fine-tune the pruned models by employing LoRA which only updates two injected low-rank decomposition matrices that are attached to a frozen pre-trained weight matrix. Given two low-rank matrices $A\in \mathbb{R}^{r \times k}$ and $B\in \mathbb{R}^{d \times r}$ ($r \ll \text{min}(d,k)$ ), the forward computation can be written as:
\begin{equation}
    f(x) = xW_{0}+xBA,
\end{equation}
where $x \in \mathbb{R}^{n\times d}$ denotes inputs. After adaption, the updated $W$ can be re-parameterized as $W=W_{0}+BA$. 

\section{Experiments}
In this section, we evaluate the performance of our MINI-LLM on three kinds of LLMs, covering a wide range of tasks. We first introduce the experimental setup, then present the main results and provide ablation studies for further analysis. 
\subsection{Experimental Setup}
\textbf{Models, datasets, and evaluation metrics}. To verify the effectiveness and versatility of our MINI-LLM, we test it over three open-source LLMs with different structures: LLaMA-7B~\cite{touvron2023llama}, BLOOM-7B~\cite{workshop2023bloom}, and OPT-6.7B~\cite{zhang2022opt}. All models undergo evaluation in a task-agnostic framework. We assess the zero-shot ability of pruned models on WikiText2~\cite{merity2016pointer} and PTB~\cite{marcus1993building} for language generation with the perplexity (PPL)\footnote{https://huggingface.co/spaces/evaluate-metric/perplexity} analysis, and smaller is better. Besides, we follow LLaMA to implement zero-shot task classification and multiple-choice on four common sense reasoning datasets: BoolQ~\cite{clark2019boolq}, PIQA~\cite{bisk2020piqa}, HellaSwag~\cite{zellers2019hellaswag}, and WinoGrande~\cite{sakaguchi2021winogrande}. In addition to zero-shot evaluation, we conduct experiments on few-shot tasks to evaluate pruned LLMs' ability to learn in context. We choose the Massive Multitask Language Understanding benchmark (\textbf{MMLU}) [\cite{hendrycks2021measuring}] and conduct a 5-shot evaluation to remain consistent with the evaluation approach described by [\cite{touvron2023llama}]. In task classification and multiple-choice on common sense reasoning datasets, as well as on MMLU, classification accuracy is used as the performance metric.

\begin{table*}[t]
    \centering
    \resizebox{\textwidth}{!}{
    \small
    \begin{tabular}{lr|r|rr|rrrrr}
        \toprule
        \midrule
        Prune ratio  & Method & GPU (GB) & WikiText2$\downarrow$ & PTB$\downarrow$  & BoolQ & PIQA & HellaSwag & WinoGrande & Average$\uparrow$\\
        \midrule
        \multirow{2}{*}{0\%} & LLaMA-7B~\cite{touvron2023llama}* & 0 & - & - & 76.50 & 79.8 & 76.10 & 70.10 & 75.63\\
        & LLaMA-7B~\cite{ma2023llmpruner}* & 0 & 12.62 & 22.15 & 73.18 & 78.35 & 72.99 & 67.01 & 72.88  \\
        \midrule
        \multirow{5}{*}{20\% w/ tune} & LLM-Pruner~\cite{ma2023llmpruner}* & 40.00 & 17.58 & 30.11 & 64.62 & 77.20 & 68.80 & 63.14 & 68.44   \\
        \cline{2-10}
        & magnitude-l1 & 21.60 & 24.32 & 43.19 & 58.47 & 75.35 & 65.40 & 60.93 & 65.04 \\
        & magnitude-l2 & 21.60 & 24.23 & 36.16 & \underline{65.02} & 75.14 & 65.07 & 62.12 & \underline{66.84} \\
        & SparseGPT~\cite{frantar2023sparsegpt} & 40.00 & 20.84 & 36.23 & 55.05 & \textbf{75.84} & \textbf{66.88} & 61.64 & 64.85\\
        & Wanda~\cite{sun2024simple} & 21.90 & \underline{20.36} & \underline{36.15} & 60.92 & 74.70 & \underline{66.70} & \underline{62.33} & 66.16 \\
        & MINI-LLM (ours) & 22.40 & \textbf{18.32} & \textbf{32.54} & \textbf{66.76} & \underline{75.46} & 65.61 & \textbf{62.43} & \textbf{67.57} \\
        \midrule
        \multirow{5}{*}{30\% w/ tune} & LLM-Pruner~\cite{ma2023llmpruner}  & 37.16 & 21.55 & 37.67 & 64.89 & 73.72 & 63.45 & 62.67 & 66.18 \\
        \cline{2-10}
        & magnitude-l1 & 20.10 & 31.17 & 54.28 & 61.80 & 73.23 & 58.02 & 56.91 & 62.49 \\
        & magnitude-l2 & 20.10 & 31.11 & 51.29 & \underline{61.89} & 73.45 & 57.84 & 58.72 & \underline{62.98} \\
        & SparseGPT~\cite{frantar2023sparsegpt} & 40.00 & \underline{26.82} & \underline{45.71} & 56.64 & 73.45 & \textbf{61.23} & \underline{59.51} & 62.71\\
        & Wanda~\cite{sun2024simple} & 20.33 & 27.08 & 46.25 & 56.97 & \textbf{74.27} & \underline{59.27} & \textbf{60.46} & 62.74 \\
        & MINI-LLM (ours) & 20.90 & \textbf{24.28} & \textbf{39.02} & \textbf{64.55} & \underline{73.74} & 58.74 & 58.93  & \textbf{63.99}  \\
        \midrule
        \multirow{5}{*}{40\% w/ tune}  & LLM-Pruner~\cite{ma2023llmpruner}  & 36.13 & 28.10 & 48.66 & 60.46 & 71.33 & 55.62 & 56.43 & 60.96 \\
        \cline{2-10}
        & magnitude-l1 & 18.80 & 43.96 & 66.63 & 47.03 & 70.89 & 49.79 & 52.96 & 55.17 \\
        & magnitude-l2 & 18.80 & 45.26 & 67.68 & 48.72 & \textbf{71.65} & 50.21 & 53.43 & 56.00 \\
        & SparseGPT~\cite{frantar2023sparsegpt} & 40.00 & 37.16 & \underline{66.12} & \underline{58.26} & 71.44 & \textbf{53.91} & \underline{56.99} & \underline{60.15} \\
        & Wanda~\cite{sun2024simple} & 19.13 & \underline{36.44} & 66.37 & 49.91 & 71.38 & \underline{53.85} & \textbf{58.72} & 58.47 \\
        & MINI-LLM (ours) & 19.83 & \textbf{31.78} & \textbf{49.23} & \textbf{63.65} & \underline{71.59} & \underline{53.31} & 55.56 & \textbf{61.02} \\  
        \midrule
        \multirow{5}{*}{50\% w/ tune}  & LLM-Pruner~\cite{ma2023llmpruner}* & 35.00 & 38.12 & 66.35 & 60.28 & 69.31 & 47.06 & 53.43 & 57.52  \\
        \cline{2-10}
        & magnitude-l1 & 17.59 & 61.39 & 91.79 & 40.73 & 66.32 & 42.66 & 51.85 & 50.39  \\
        & magnitude-l2 & 17.59 & 58.12 & 89.67 & 38.50 & 67.08 & 43.47 & 52.80 & 50.46 \\
        & SparseGPT~\cite{frantar2023sparsegpt} & 40.00 & 49.52 & 82.28 & \underline{42.84} & \underline{68.01} & \underline{45.38} & \textbf{55.96} & \underline{53.05} \\
        & Wanda~\cite{sun2024simple} & 18.82 & \underline{45.98} & \underline{78.82} & 40.49 & \textbf{69.04} & 45.10 & \underline{54.93} & 52.39 \\
        & MINI-LLM (ours) & 18.82 & \textbf{44.69} & \textbf{69.83} & \textbf{61.35} & 67.85 & \textbf{45.39} & 53.12 & \textbf{56.93} \\ 
        \midrule
        \bottomrule
    \end{tabular}}
    \vspace{-1mm}
    \caption{Zero-shot performance of the pruned LLaMA-7B models. ``Prune Ratio'' refers to the proportion of parameters removed relative to the original number of parameters. ``GPU (GB)'' indicates the peak GPU memory usage for pruning. ``Average'' is calculated among four classification datasets. \textbf{Bold}/\underline{Underline} mark the best/second best performance at the same compression rate with fine-tuning, respectively, excluding LLM-Pruner in the comparison. An asterisk (*) signifies the results are taken directly from the corresponding papers.}
    \label{tab:llama-7b-oneshot}
    \vspace{-1mm}    
    \flushleft{\footnotesize $^{1}$We have saved the scores of each weight for SparseGPT, which results in high GPU memory usage. This can be optimized through implementation.}
    \vspace{-2mm}
\end{table*}

\vspace{0.2cm}
\noindent \textbf{Pruning and fine-tuning settings}. Our MINI-LLM conducts in a one-shot pruning framework. That is scoring only once and then pruning the network to the target prune ratio~\cite{cheng2023survey}. In the model pruning process, we use 10 randomly selected samples from Bookcorpus~\cite{zhu2015aligning} as the calibration data for evaluating the weight gradients and 128 samples for computing each layer's input (i.e., activation). Due to the varying sensitivity of each layer to pruning~\cite{ma2023llmpruner}, the first four layers and the last three layers are retained. During the recovery phase, we utilize the Alpaca-cleaned~\cite{taori2023stanford} as the training dataset, which contains approximately 50k samples, to fine-tune the pruned models with a batch size of 64. Following [\cite{ma2023llmpruner}], the learning rate is set to 1e-4 and a total of 2 epochs. Each pruned model is recovered by an Adam optimizer~\cite{kingma2015adam} paired with a cosine decay schedule for the learning rate. We set LoRA $r=8$, $\alpha=16$, and attach LoRA modules on all linear layers of the base model. In the inference stage, all the evaluations are implemented with a context length of 128.

\noindent \textbf{Baselines}. We compare MINI-LLM with four one-shot structured pruning methods for LLMs. Magnitude-l1/l2: pruning based on the absolute values or the l2-norm of weights, respectively. LLM-Pruner~\cite{ma2023llmpruner}: pruning using criterion Eq.~\eqref{eq:sensitivity-score-taylor} with backpropagation gradients. Wanda~\cite{sun2024simple}: pruning based on the product of the magnitude of weights and their corresponding activations. Given that vanilla SparseGPT and Wanda are retraining-free unstructured methods, we adapt them for structured pruning with pruning and fine-tuning stages for a fair comparison while maintaining the same criterion. Except for LLM-pruner, which is a gradient-based method, the other methods are all gradient-free methods.

\subsection{Main Results}
\textbf{Zero-shot performance on LLaMA-7B.} 
We prune LLaMA-7B with four prune ratios: from 20\% to 50\% and fine-tune the pruned models by using LoRA to restore model accuracy. The comparisons with the baselines are reported in Table~\ref{tab:llama-7b-oneshot}. From the results, we see that our MINI-LLM consistently surpasses all the gradient-free methods and closely matches or even outperforms the LLM-Pruner with backpropagation gradients across four prune ratios. For example, at a 20\% prune ratio, MINI-LLM achieves an average classification accuracy of 67.57\% across four inference datasets, better than other gradient-free methods, and obtains 92.71\% of the accuracy achieved by the original model. Although LLM-Pruner achieves better accuracy with 93.91\% of the accuracy attained by the dense model, the peak GPU memory required for pruning by LLM-Pruner is approximately twice that of MINI-LLM. Moreover, although both Wanda and MINI-LLM use weight magnitude and activation, MINI-LLM performs better, which indicates that estimated gradients are beneficial in guiding pruning. In addition, at a 40\% prune ratio, MINI-LLM achieves an average accuracy of 61.02\% on the four tasks, even better than LLM-Pruner's average accuracy of 60.96\%. 

However, similar to the observation in Ma et al.~\shortcite{ma2023llmpruner}, with a high prune ratio, such as 50\%, an obvious performance decline is observed, as shown in Table~\ref{tab:llama-7b-oneshot}. In this situation, our MINI-LLM and LLM-Pruner only retain 78.11\% and 78.92\% of the dense model's accuracy, respectively. Even for LLMs, structurally pruning under high prune ratios remains a major challenge.

\vspace{0.2cm}
\noindent \textbf{Zero-shot performance on BLOOM-7B and OPT-6.7B.} To validate MINI-LLM on other LLMs broadly, we prune both BLOOM-7B and OPT-6.7B with two prune ratios: 10\% and 30\%, and fine-tune the pruned models to restore model accuracy. The results in Table~\ref{tab:bloom-opt-oneshot} illustrate that our MINI-LLM steadily outperforms all gradient-free methods and exhibits performance comparable to, even surpasses at times, that of LLM-Pruner. For instance, at a 30\% compression rate on BLOOM-7B, MINI-LLM achieves a perplexity of 54.07 on the WikiText2 dataset, obviously outperforming LLM-Pruner's perplexity of 58.11. Similarly, at a 30\% compression rate on OPT-6.7B, MINI-LLM achieves a perplexity of 40.89 on the WikiText2 dataset and 57.44 on the PTB dataset, outperforming LLM-Pruner's perplexity of 42.94 and 65.09, respectively. In addition, at a 10\% prune ratio on OPT-6.7B, MINI-LLM achieves an average classification accuracy of 67.81\% across four datasets and obtains 98.60\% of the accuracy achieved by the original model, which is even better than LLM-Pruner' s 67.50\% and 98.15\%. This demonstration validates the effectiveness of MINI-LLM in efficiently compressing models of various structures to a specified size, while optimizing memory usage.

\begin{table*}[t]
    \centering
    \resizebox{\textwidth}{!}{
    \small
    \begin{tabular}{lr|r|rr|rrrrr}
        \toprule
        \midrule
        Prune ratio  & Method & GPU (GB) & WikiText2$\downarrow$ & PTB$\downarrow$  & BoolQ & PIQA & HellaSwag & WinoGrande & Average$\uparrow$\\
        \midrule
        0\% & BLOOM-7B~\cite{workshop2023bloom} & 0 & 26.58 & 50.55 & 62.94 & 73.61 & 59.69 & 64.4 & 65.16\\
        \midrule
        \multirow{5}{*}{10\% w/ tune} & LLM-Pruner~\cite{ma2023llmpruner} & 40.00 & 35.32 & 75.21 & 62.14 & 72.14 & 55.39 & 57.85 & 61.88   \\
        \cline{2-10}
        & magnitude-l1 & 22.04 & 40.89 & 92.87 & 59.28 & 71.82 & 52.44 & 56.21 & 59.94 \\
        & magnitude-l2 & 22.04 & 40.73 & 95.45 & 59.33 & 72.04 & 52.58 & 56.04 & 60.00 \\
        & SparseGPT~\cite{frantar2023sparsegpt} & 40.00 & \underline{40.42} & \underline{92.15} & 59.17 & 70.73 & 52.38 & 56.2 & 59.62\\
        & Wanda~\cite{sun2024simple} & 22.81 & 40.81 & 93.60 & \underline{59.94} & \textbf{72.25} & \underline{52.60} & \textbf{57.14} & \underline{60.48} \\
        & MINI-LLM (ours) & 25.03 & \textbf{38.12} & \textbf{86.23} & \textbf{59.97} & \underline{72.05} & \textbf{53.54} & \underline{56.43} & \textbf{60.50} \\
        \midrule
        \multirow{5}{*}{30\% w/ tune} & LLM-Pruner~\cite{ma2023llmpruner}  & 38.51 & 58.11 & 147.52 & 62.11 & 67.79 & 44.04 & 53.28 & 56.81 \\
        \cline{2-10}
        & magnitude-l1 & 19.49 & 87.25 & \underline{166.21} & \underline{61.04} & 65.40 & 41.46 & 51.70 & \underline{54.90} \\
        & magnitude-l2 & 19.49 & 79.75 & 167.83 & 59.45 & 66.87 & 42.36 & 50.91 & 54.89 \\
        & SparseGPT~\cite{frantar2023sparsegpt} & 40.00 & \underline{75.51} & 173.51 & 52.02 & \underline{67.14} & \underline{42.86} & \textbf{53.28} & 53.83\\
        & Wanda~\cite{sun2024simple} & 20.34 & 84.89 & 170.16 & 53.61 & 67.03 & 41.34 & \underline{50.99} & 53.24 \\
        & MINI-LLM (ours) & 22.11 & \textbf{54.07} & \textbf{121.61} & \textbf{62.17} & \textbf{68.82} & \textbf{44.95} & \underline{51.93} & \textbf{56.97}  \\
        \midrule
        0\% & OPT-6.7B~\cite{zhang2022opt} & 0 & 26.45 & 32.03 & 66.06 & 76.55 & 67.21 & 65.27 & 68.77  \\
        \midrule
        \multirow{5}{*}{10\% w/ tune}  & LLM-Pruner~\cite{ma2023llmpruner}  & 38.00 & 27.89 & 39.33 & 63.06 & 76.77 & 66.33 & 63.85 & 67.50 \\
        \cline{2-10}
        & magnitude-l1 & 22.81 & 39.17 & 55.68 & 58.17 & 75.30 & 60.35 & 59.59 & 63.35 \\
        & magnitude-l2 & 22.81 & 39.40 & 54.49 & 59.08 & 74.86 & 60.45 & 60.22 & 63.65 \\
        & SparseGPT~\cite{frantar2023sparsegpt} & 40.00 & \underline{36.58} & \underline{50.99} & 61.93 & 74.81 & 61.25 & 60.46 & 64.61\\
        & Wanda~\cite{sun2024simple} & 23.04 & 37.09 & 53.54 & \textbf{66.09} & \underline{75.46} & \underline{62.24} & \underline{62.59} & \underline{66.60} \\
        & MINI-LLM (ours) & 23.65 & \textbf{30.15} & \textbf{38.64} & \underline{65.90} & \textbf{76.12} & \textbf{65.66} & \textbf{63.54} & \textbf{67.81} \\  
        \midrule
        \multirow{5}{*}{30\% w/ tune}  & LLM-Pruner~\cite{ma2023llmpruner} & 34.66 & 42.94 & 65.09 & 61.93 & 73.83 & 56.98 & 59.98 & 63.60\\
        \cline{2-10}
        & magnitude-l1 & 19.91 & 81.96 & 104.01 & 48.50 & \underline{69.59} & \underline{44.99} & 53.75 & 54.21  \\
        & magnitude-l2 & 19.91 & \underline{76.10} & \underline{98.86} & 54.22 & 69.37 & 44.83 & 54.14 & 56.11 \\
        & SparseGPT~\cite{frantar2023sparsegpt} & 40.00 & 77.00 & 103.61 & 54.31 & 69.21 & 44.56 & \underline{55.56} & 55.91\\
        & Wanda~\cite{sun2024simple} & 20.07 & 82.93 & 107.32 & \underline{60.34} & 69.53 & 44.58 & 54.46 & \underline{57.23} \\
        & MINI-LLM (ours) & 20.65 & \textbf{40.89} & \textbf{57.44} & \textbf{62.17} & \textbf{72.58} & \textbf{54.07} & \textbf{56.20} & \textbf{61.26} \\ 
        \midrule
        \bottomrule
    \end{tabular}}
    \vspace{-1mm}
    \caption{Zero-shot performance of the pruned BLOOM-7B and OPT-6.7B. Columns is consistent with the definitions in Table~\ref{tab:llama-7b-oneshot}. Unless otherwise specified, ``Prune Ratio'' and \textbf{Bold}/\underline{Underline} have the same meaning as Table~\ref{tab:llama-7b-oneshot}.}
    \label{tab:bloom-opt-oneshot}
    \vspace{-2mm}
\end{table*}

\begin{figure}[t]
\centering
  \subfloat[BLOOM-7B]{
  \vspace{-2mm}
  \begin{minipage}{4cm}
  \includegraphics[width=4cm,height=3cm]{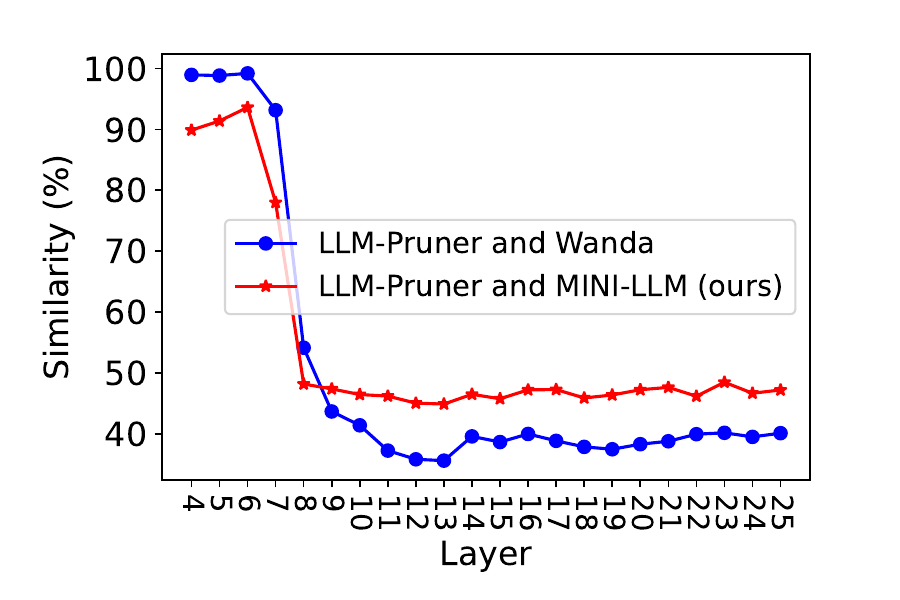}
  \label{Fig:similarity-bloom-7b1-ratio-0.3}
  \end{minipage}
  }
  \subfloat[OPT-6.7B]{  
  \begin{minipage}{4cm}
  \vspace{-2mm}
  \includegraphics[width=4cm,height=3cm]{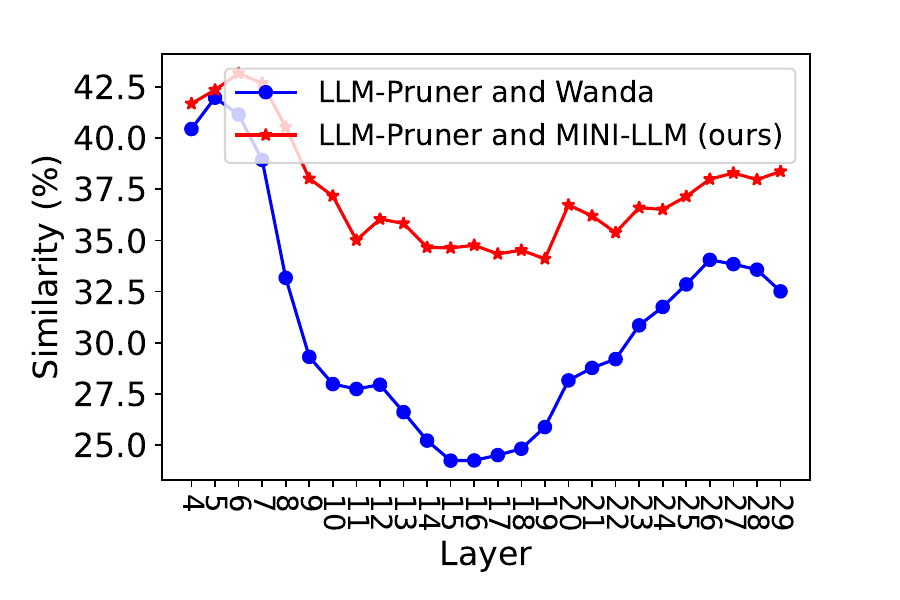}
  \label{Fig:similarity-opt-6.7b-ratio-0.3}
  \end{minipage} 
  }
 \vspace{-1mm}
 \caption{Similarity in pruned channels at the prune ratio of 30\%. LLM-Pruner and MINI-LLM (ours) have more similar pruned channels compared to LLM-Pruner and Wanda.}
 \label{Fig:similarity-channels}
\vspace{-0.2cm}
\end{figure}

In addition, we observe that the pruning outcomes achieved by gradient-free methods such as Wanda and magnitude l1/l2 shown in Table~\ref{tab:bloom-opt-oneshot} significantly fell short in comparison to gradient-based pruning methods such as LLM-Pruner and MINI-LLM at a prune ratio of 30\% on the WikiText2 and PTB datasets for BLOOM and OPT. Using LLM-Pruner as a high-quality benchmark, we compare Wanda, representing gradient-free approaches, by assessing the similarity of their retained channels per layer against LLM-Pruner on the WikiText2 dataset. Similarly, we evaluate the similarity between LLM-Pruner and MINI-LLM. Specifically, the similarity is calculated by the formula: $\lvert | \text{Intersection}(A, B) \rvert|_{0}/\lvert | A \rvert |_{0} \times 100\%$, where $A$ and $B$ denote the sets of the pruned channels obtained by LLM-Pruner and the examined method, respectively.
The results are illustrated in Figure~\ref{Fig:similarity-channels}. We can see that LLM-Pruner and MINI-LLM have more similar pruned channels compared to LLM-Pruner and Wanda. As a result, compared to gradient-free methods, the perplexity of MINI-LLM in Table~\ref{tab:bloom-opt-oneshot} is closer to the results of LLM-Pruner.

\begin{figure}[t]
\centering
  \subfloat[WikiText2]{
  \begin{minipage}{4cm}
  \includegraphics[width=4cm,height=3cm]{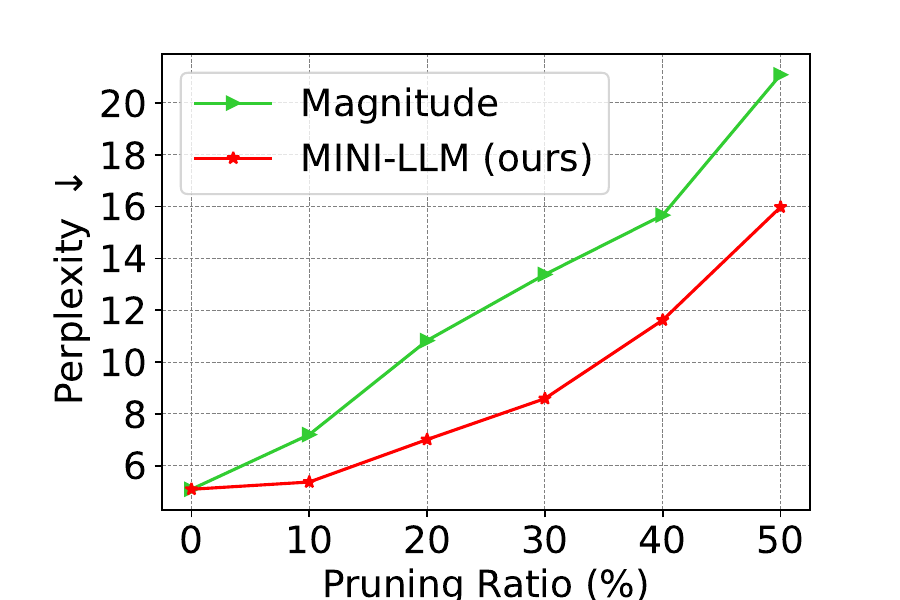}
     \label{Fig:llama-13b-wikitext2}
  \end{minipage}
  }
  \subfloat[PTB]{  
  \begin{minipage}{4cm}
  \includegraphics[width=4cm,height=3cm]{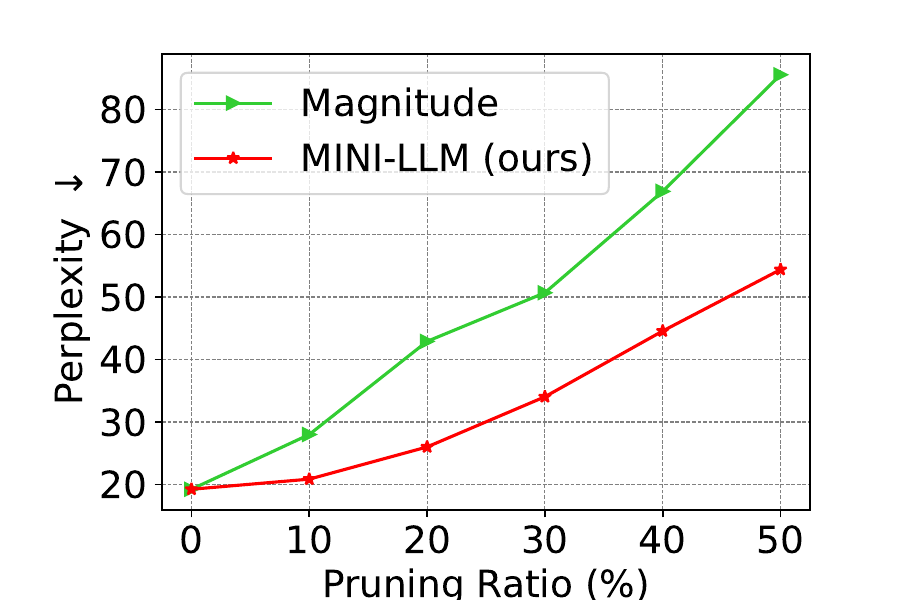}
     \label{Fig:llama-13b-ptb}
  \end{minipage} 
  }  
 \caption{Zero-shot perplexity of the pruned LLaMA-13B models when fine-tuning is applied. MINI-LLM consistently maintains its substantial advantage over the magnitude-based method across a spectrum of pruning ratios.}
 \label{Fig:llama-13b}
\end{figure}

\vspace{0.2cm}
\noindent \textbf{Zero-shot Performance on LLaMA-13B}
Due to the efficient approximation for the gradients of the pre-trained weights, MINI-LLM enables pruning on larger-scale LLMs, such as LLaMA-13B~\footnote{https://huggingface.co/huggyllama/llama-13b/tree/main}. We prune LLaMA-13B with five pruning ratios: from 10\% to 50\% and present the zero-shot performance of the pruned LLaMA-13B without fine-tuning in Table~\ref{tab:llama-13b-oneshot} and with fine-tuning in Figure~\ref{Fig:llama-13b}, respectively. Except for the model pruned with a ratio of 50\%, for which fine-tuning is conducted for two epochs, the compressed models obtained from all other pruning ratios are fine-tuned for just one epoch. The other fine-tuning settings, such as the learning rate and batch size, are the same as those for recovering LLaMA-7B. We follow Wanda~\cite{sun2024simple} to conduct inference with 2048 tokens. 

As shown in Table~\ref{tab:llama-13b-oneshot}, MINI-LLM outperforms the magnitude-based method (l2-norm) significantly when fine-tuning is not applied. For example, with a pruning ratio of 30\%, MINI-LLM achieves a perplexity of 11.04 compared to the magnitude-based method's 316.65 on the WikiText2 dataset. Similarly, as depicted in Figure~\ref{Fig:llama-13b}, MINI-LLM consistently maintains its substantial advantage over the magnitude-based method across a spectrum of pruning ratios when subjected to fine-tuning.  

\begin{table*}[t]
    \centering
    \small
    \scalebox{0.80}{%
    \begin{tabular}{l|r|rrrrrr}
    \toprule
    \midrule    
    \multirow{2}{*}{Method} & \multirow{2}{*}{Dataset} & \multicolumn{4}{r}{Prune Ratio} \\
     \cline{3-8}
     &  & 0\% & 10\% & 20\% & 30\% & 40\% & 50\% \\ 
     \midrule
     Magnitude (l2-norm) & \multirow{2}{*}{WikiText2} & \textbf{5.09} & 13.78 & 21.42 & 316.65 & 3918.40 & 12550.30\\
     MINI-LLM &  & \textbf{5.09} & \textbf{5.87} & \textbf{7.51} & \textbf{11.04} & \textbf{22.32} & \textbf{115.13}\\
     \midrule
     Magnitude (l2-norm) & \multirow{2}{*}{PTB} & \textbf{19.24} & 50.89 & 91.87 & 694.34 & 4236.80 & 12847.92\\
     \cline{3-8}
     MINI-LLM &  & \textbf{19.24} & \textbf{23.86} & \textbf{33.38} & \textbf{49.91} & \textbf{91.43} & \textbf{176.93} \\
    \midrule
    \bottomrule
    \end{tabular}
    }
   \vspace{-1mm}    
   \caption{Zero-shot perplexity of the pruned LLaMA-13B when fine-tuning is not applied.}
   \vspace{-2mm}   
   \label{tab:llama-13b-oneshot}
\end{table*}

\vspace{0.2cm}
\noindent \textbf{Few-shot performance on LLaMA-7B.} 
In Table~\ref{tab:few-shot-for-pruned-llama}, we report the mean accuracies for both dense LLMs and sparse LLMs with 20\% to 50\% sparsity. In the few-shot setting, MINI-LLM performs competitively with other methods, including backpropagation gradient-based LLM-Pruner. Specifically, at a 20\% prune ratio, MINI-LLM achieves an average accuracy of 26.60\%, which surpasses SparseGPT's 25.80\% and LLM-Pruner's 25.30\%. Notably, estimated gradient-based MINI-LLM consistently surpasses backpropagation gradient-based LLM-Pruner. This performance is not observed in zero-shot setting.

\begin{table}[!ht]
    \centering
    \scalebox{0.80}{%
    \small
    \begin{tabular}{lr|rrrrr}
    \toprule
    \midrule
    \multirow{2}{*}{Ratio} & \multirow{2}{*}{Method} &  \multicolumn{5}{c}{MMLU (5-shot)} \\
    & & STEM & Humans & Social & Other & Avg. \\
    \midrule 
    0\% & - & 32.60 & 34.10 & 40.40 & 40.90 & 36.70\\
    \midrule
    \multirow{3}{*}{20\%} & SparseGPT~\shortcite{frantar2023sparsegpt} & \underline{25.30} & \textbf{25.90} & \underline{25.50} & 26.30 & \underline{25.80} \\
    & Wanda~\shortcite{sun2024simple} & 23.30 & \underline{25.80} & 23.20 & 24.60 & 24.40\\
    & LLM-Pruner~\shortcite{ma2023llmpruner} & 24.40 & 25.30 & 23.80 & \underline{27.30} & 25.30\\
    & MINI-LLM (ours) & \textbf{25.50} & \textbf{25.90} & \textbf{26.20} & \textbf{29.00} & \textbf{26.60}\\
    \midrule
    \multirow{3}{*}{30\%} & SparseGPT~\shortcite{frantar2023sparsegpt} & \textbf{25.80} & \underline{25.70} & 24.60 & 23.50 & 25.00 \\
    & Wanda~\shortcite{sun2024simple} & \textbf{25.80} & \textbf{27.10} & \underline{25.00} & 24.80 & \textbf{25.80}\\
    & LLM-Pruner~\shortcite{ma2023llmpruner} & 23.90 & 24.90 & 23.50 & \underline{26.00} & 24.60\\
    & MINI-LLM (ours) & \underline{24.10} & 24.80 & \textbf{25.70} & \textbf{26.20} & \underline{25.20}\\
    \midrule
     \multirow{3}{*}{40\%} & SparseGPT~\shortcite{frantar2023sparsegpt} & \underline{26.10} & \textbf{25.50} & 23.30 & 23.90 & 24.80 \\
    & Wanda~\shortcite{sun2024simple} & 25.80 & \underline{24.50} & \underline{24.80} & 23.60 & \underline{25.80}\\
    & LLM-Pruner~\shortcite{ma2023llmpruner} & 22.70 & 24.40 & 21.40 & \underline{24.00} & 23.30\\
    & MINI-LLM (ours) & \textbf{26.20} & 24.10 & \textbf{27.50} & \textbf{27.50} & \textbf{26.10}\\
    \midrule
    \multirow{3}{*}{50\%} & SparseGPT~\shortcite{frantar2023sparsegpt} & \textbf{26.40} & 24.70 & \textbf{25.40} & 24.20 & \underline{25.10} \\
    & Wanda~\shortcite{sun2024simple} & 26.00 & \textbf{25.10} & 24.30 & \underline{25.00} & \underline{25.10}\\
    & LLM-Pruner~\shortcite{ma2023llmpruner} & 21.30 & 24.20 & 21.70 & 23.70 & 22.90\\
    & MINI-LLM (ours) & \underline{26.30} & \underline{24.80} & \underline{25.00} & \textbf{25.30} & \textbf{25.30}\\
    \midrule
    \bottomrule
    \end{tabular}
    }
    \vspace{-1mm}
    \caption{Few-shot performance of the pruned LLaMA-7B models. ``Ratio'', \textbf{Bold}/\underline{Underline}, ``Avg.'' have the same meaning as Table~\ref{tab:llama-7b-oneshot}}
    \label{tab:few-shot-for-pruned-llama}
    \vspace{-2mm}
\end{table}

\vspace{0.2cm}
\noindent \textbf{Model size, complexity, and inference time.}  
Table~\ref{tab:statics-for-pruned-llama} shows the number of parameters, MACs, GPU memory requirements, and total inference time for running the original model and the pruned LLaMA-7B models at different prune ratios. The results indicate that when the model is pruned by 50\%, the total inference time is reduced to 58\% and the GPU memory usage concurrently drops to 50\% of its original values, respectively. The evaluation is conducted in the inference mode and the sequence length is set to 64. The inference time is tested under the test dataset of WikiText2 on a single NVIDIA GeForce RTX 3090Ti (24GB).

\subsection{Ablation Study}
\begin{figure}[t]
  \centering
  \vspace{-3mm}
  \subfloat[WikiText2]{
  \begin{minipage}{4cm}
  \includegraphics[width=4cm,height=3cm]{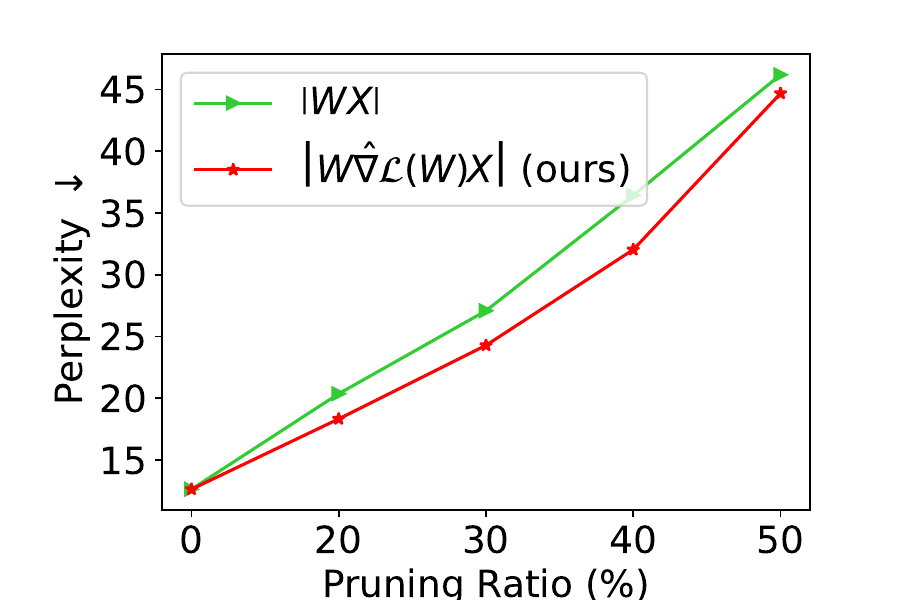}
     \label{Fig:ppl-with-without-activation-wiki}
  \end{minipage}
  }
  \subfloat[PTB]{  
  \begin{minipage}{4cm}
  \includegraphics[width=4cm,height=3cm]{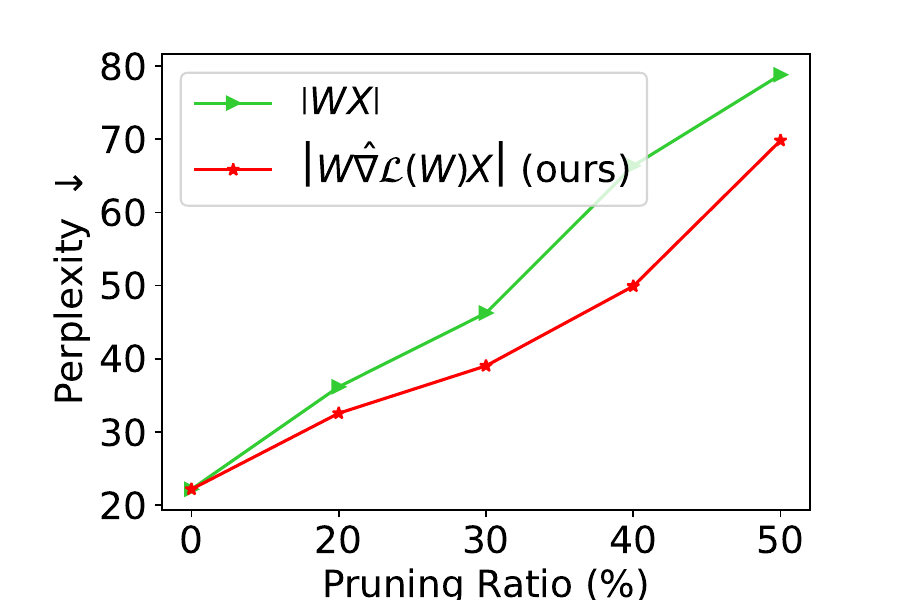}
     \label{Fig:ppl-with-without-activation-ptb}
  \end{minipage} 
  }
 \vspace{-1mm} 
 \caption{The outcomes of gradient-based vs. gradient-free criteria for pruning LLaMA-7B. The results demonstrate that our score function FMS consistently yields better performance compared to Wanda’s pruning criterion.}
 \label{Fig:gradient-with-without-activation}
\vspace{-0.2cm}
\end{figure}

\textbf{Efficacy of estimated gradients on LLaMA-7B.} To enhance GPU memory efficiency over traditional backpropagation gradients, we utilize the classical ZO gradient estimation based on SPSA to approximately compute weight gradients with only forward passes for LLM pruning. Although SPSA-based ZO optimization is theoretically founded (\cite{spall1992multivariate,spall1997onemeasurement,gasnikov2022randomized}), we especially reveal the effectiveness of the estimated gradients for guiding pruning LLMs in Figure~\ref{Fig:gradient-with-without-activation}. As we can see, the results demonstrate that our score function FMS, $\left | W\hat{\nabla}\mathcal{L}(W)X \right |$ (represented by the red line in Figure~\ref{Fig:ppl-with-without-activation-wiki}), consistently yields better performance compared to Wanda's pruning criterion, $\left | WX \right |$ (indicated by the green line), across four prune ratios ranging from 20\% to 50\% for LLaMA-7B on the WikiText2 dataset. On the PTB dataset, this improved performance is more evident, as shown in Fig~\ref{Fig:ppl-with-without-activation-ptb}. Comparing these two criteria, our FMS includes an additional estimated gradient $\hat{\nabla}\mathcal{L}(W)$ compared to Wanda's. This indicates that the performance improvement over Wanda's comes from the estimated gradient information. This observation underscores the superior performance and effectiveness of gradient-based pruning methods in our experiments. However, as we previously mentioned, gradients based on backpropagation lead to substantial memory consumption and are less feasible. Therefore, gradient estimation based on the forward passes becomes valuable, allowing the criterion to incorporate guidance information from gradients. 

\begin{table}[!ht]
    \centering
    \scalebox{0.95}{%
    \small
    \begin{tabular}{l|rrrr}
    \toprule
    \midrule
    Ratio & \#Params & \#MACs & GPU Memory  &  Inference Time \\ 
    \midrule 
    0\% & 6.74B & 424.02G & 12884.5MB & 88.81s \\
    20\% & 5.42B & 340.48G & 10375.5MB & 71.77s\\
    50\% & 3.39B & 279.37G & 6519.0MB & 51.18s\\ 
    \midrule
    \bottomrule
    \end{tabular}
    }
    \vspace{-1mm}
    \caption{Model size, complexity, and inference time of the original model and the pruned LLaMA-7B models. ``Inference Time'' means the total inference time on WikiText2 test dataset. The evaluation is conducted in inference mode with a sequence length of 64, and the inference time is tested on a single NVIDIA GeForce RTX 3090 Ti (24GB).}
    \label{tab:statics-for-pruned-llama}
    \vspace{-2mm}
\end{table}

\vspace{0.2cm}
\noindent \textbf{Efficacy of Estimated Gradients on BLOOM-7B and OPT-6.7B}
In the main body, we explored the effectiveness of the estimated gradients for guiding the pruning of LLaMA-7B. Here, we delve further into the effectiveness of the estimated gradients in guiding the pruning process for BLOOM-7B~\footnote{https://huggingface.co/bigscience/bloom-7b1/tree/main} and OPT-6.7B~\footnote{https://huggingface.co/facebook/opt-6.7b/tree/main}. The experimental results presented in Table~\ref{tab:gradient-based-gradient-free-bloom-opt} indicate that our score function FMS, $| W\hat{\nabla}\mathcal{L}(W)X |$, consistently yields better performance compared to Wanda's pruning criterion, $| WX |$, for pruning BLOOM-7B and OPT-6.7B on the WikiText2 and PTB datasets. For example, at a 30\% pruning ratio, $| W\hat{\nabla}\mathcal{L}(W)X |$ achieves a perplexity of 54.07 on WikiText2 for pruning BLOOM-7B. This result shows a 30.82 improvement over $| WX |$. Similarly, at a 30\% pruning ratio, $| W\hat{\nabla}\mathcal{L}(W)X |$ achieves a perplexity of 40.89 on WikiText2 for pruning OPT-6.7B, which surpasses the perplexity of 82.93 obtained by using $| WX |$. 

\begin{table}[t]
    \centering
    \small
    \scalebox{0.90}{%
    \begin{tabular}{l|r|rr|rr}
    \toprule
    \midrule    
    Prune Ratio & Model/Criterion & WikiText2$\downarrow$ & $\Delta \uparrow$  & PTB$\downarrow$ & $\Delta \uparrow$ \\ 
    \midrule
    0\% & 
    BLOOM-7B & 26.58 & - & 50.55 & -\\
    \midrule
    10\% & $| W X |$ & 40.81 & \multirow{2}{*}{2.69} & 93.60 & \multirow{2}{*}{7.37}\\
    w/ tune & $| W \hat{\nabla} \mathcal{L}(W) X |$ & \textbf{38.12} & & \textbf{86.23} & \\
    \midrule 
    30\% & 
    $| W X |$ & 84.89 & \multirow{2}{*}{30.82} & 170.16 & \multirow{2}{*}{48.55}\\
    w/ tune & $| W \hat{\nabla}\mathcal{L}(W) X|$ & \textbf{54.07} & & \textbf{121.61} & \\
    \midrule
    0\% & 
    OPT-6.7B & 26.45 & - & 32.03 & -\\
    \midrule
    10\% & 
    $| W X |$ & 37.09 & \multirow{2}{*}{6.94} & 53.54 & \multirow{2}{*}{14.90}\\
    w/ tune & $| W \hat{\nabla} \mathcal{L}(W) X |$ & \textbf{30.15} & & \textbf{38.64} & \\
    \midrule 
    30\% & 
    $| W X |$ & 82.93 & \multirow{2}{*}{42.04} & 107.32 & \multirow{2}{*}{49.88}\\
    w/ tune & $| W \hat{\nabla}\mathcal{L}(W) X|$ & \textbf{40.89} & & \textbf{57.44} & \\
    \midrule    
    \bottomrule
    \end{tabular}
    }
   \caption{The outcomes of gradient-based vs. gradient-free criteria for pruning BLOOM-7B and OPT-6.7B. $\Delta$ represents the difference in perplexity between the pruning criterion without gradients and the one with gradients. A larger $\Delta$ value indicates a greater improvement in performance.}
    \label{tab:gradient-based-gradient-free-bloom-opt}
\end{table}

\vspace{0.2cm}
\noindent \textbf{Efficacy of activation on LLaMA-7B.} 
\cite{dettmers2022llmint8,kovaleva2021bertbusters} identified a distinct property of LLMs that a few hidden state features possess notably high magnitudes. Eliminating these features results in a considerable decline in performance. As argued in Section~\ref{sec:our-pruning-criterion}, the vanilla pruning criterion $\left | W\nabla \mathcal{L}(W) \right |$ ( Eq.~\eqref{eq:sensitivity-score-taylor}) does not highlight this characteristic of LLMs. To validate that activation in FMS, i.e., $\left | W\nabla \mathcal{L}(W) X \right |$ (Eq.~\eqref{eq:sensitivity-score-ours}), can bring improved performance, we compare the performance of the pruned models obtained by using the criteria Eq.~\eqref{eq:sensitivity-score-taylor} and Eq.~\eqref{eq:sensitivity-score-ours} over four prune ratios on the LLM. The difference between these two criteria is that Eq.~\eqref{eq:sensitivity-score-ours} includes an additional activation term $X$ compared to Eq.~\eqref{eq:sensitivity-score-taylor}. From the results shown in Table~\ref{tab:taylor-with-vs-without-activations}, we can see that the activations enable an effective increase in performance on both the WikiText2 and PTB datasets. For example, at a 20\% prune ratio, the model pruned with $\left | W\nabla \mathcal{L}(W) X\right |$ achieves a perplexity of 17.45 on the WikiText2 dataset. This result surpasses the 17.79 perplexity achieved by the model compressed by using $\left | W\nabla \mathcal{L}(W)\right |$. In contrast, on the PTB dataset, the performance enhancement provided by activation is generally more noticeable than on WikiText2. For instance, with 50\% parameters pruned, the model pruned with $\left | W\nabla \mathcal{L}(W) X\right |$ achieves a perplexity of 62.84 on the PTB dataset. Compared to using $\left | W\nabla \mathcal{L}(W)\right |$, the perplexity result decreased by 3.53. It is worth noting that activations can be estimated using a small set of calibration data and executed in a single forward pass. Therefore, computing $\left | W\nabla \mathcal{L}(W) X\right |$, as opposed to $\left | W\nabla \mathcal{L}(W)\right |$, almost does not require additional GPU memory overhead.

\begin{table}[t]
    \centering
    \small
    \scalebox{0.79}{%
    \begin{tabular}{l|r|rr|rr|rr}
    \toprule
    \midrule    
    Ratio & Criterion & WikiText2$\downarrow$ & $\Delta \uparrow$  & PTB$\downarrow$ & $\Delta \uparrow$ & Avg. $\uparrow$ & $\Delta \uparrow$ \\ 
    \midrule 
    \multirow{2}{*}{20\%} & 
    $\left | W \nabla \mathcal{L}(W) \right |$ & 17.79 & \multirow{2}{*}{0.34} & \textbf{30.57} & \multirow{2}{*}{-0.12} & 68.44 & \multirow{2}{*}{0.91} \\
    & $\left | W \nabla \mathcal{L}(W) X \right |$ & \textbf{17.45} & & 30.69 & & \textbf{69.35} & \\
    \midrule 
    \multirow{2}{*}{30\%} & 
    $\left | W \nabla\mathcal{L}(W) \right |$ & 21.55 & \multirow{2}{*}{0.38} & 37.67 & \multirow{2}{*}{0.80} & 66.18 & \multirow{2}{*}{0.39}\\
    & $\left | W \nabla\mathcal{L}(W) X \right |$ & \textbf{21.17} & & \textbf{36.87} & & \textbf{66.57} & \\
    \midrule 
    \multirow{2}{*}{40\%} & 
    $\left | W \nabla \mathcal{L}(W) \right |$ & 28.10 & \multirow{2}{*}{0.08} & 48.66 & \multirow{2}{*}{2.05} & 60.96 & \multirow{2}{*}{1.11}\\
    & $\left | W \nabla \mathcal{L}(W) X\right |$ & \textbf{28.02} & & \textbf{46.61} & & \textbf{62.07} &\\
    \midrule 
    \multirow{2}{*}{50\%} & 
    $\left |W \nabla\mathcal{L}(W) \right |$ & 39.48 & \multirow{2}{*}{0.38} & 66.37 & \multirow{2}{*}{3.53} & 57.52 & \multirow{2}{*}{0.94}\\
    & $\left | W \nabla\mathcal{L}(W) X \right |$ & \textbf{39.10} & & \textbf{62.84} & & \textbf{58.46} & \\
    \midrule    
    \bottomrule
    \end{tabular}
    }
   \vspace{-1mm}
   \caption{Zero-shot performance of the pruned LLaMA-7B models achieved by using backpropagation gradient-based pruning criterion with/without activations. ``Ratio'' refers to the prune ratio. $\Delta$ represents the difference in performance between the pruning criterion without activation and the one with activation. A larger $\Delta$ value indicates a greater improvement in performance. ``Avg.'' has the same meaning as ``Average'' in Table~\ref{tab:llama-7b-oneshot}.}
   \label{tab:taylor-with-vs-without-activations}
   \vspace{-2mm}
\end{table}

In Table~\ref{tab:taylor-with-vs-without-activations}, gradients in both pruning criteria are calculated by backpropagation. In contrast, the results in Table~\ref{tab:fwdgrad-with-vs-without-activations} demonstrate the effectiveness of activation in estimated gradients-based pruning criteria. Similar to Table~\ref{tab:taylor-with-vs-without-activations}, the difference between the two criteria in Table~\ref{tab:fwdgrad-with-vs-without-activations} also lies in whether they include activation information or not. The results in Table~\ref{tab:fwdgrad-with-vs-without-activations} indicate that using the pruning criterion with activation consistently yielded better results. For example, on the WikiText2 dataset, the model pruned with 50\% of its parameters using $|W \hat{\nabla} \mathcal{L}(W) X|$ achieves a perplexity of 44.69, reflecting a 8.54 improvement over the 53.23 perplexity by the model using $|W \hat{\nabla} \mathcal{L}(W)|$. Switching to the PTB dataset on the same compression rate, also yielded positive results, with the model's perplexity dropping from 75.50 to 69.83, confirming the efficacy of the activation-inclusive pruning criterion across diverse data.

Consequently, the results in Table~\ref{tab:taylor-with-vs-without-activations} and Table~\ref{tab:fwdgrad-with-vs-without-activations} demonstrate our pruning criterion FWS overall outperforms its counterpart without activation, whether the gradients are backpropagated or approximated.    

\begin{table}[t]
    \centering
    \small
    \scalebox{0.8}{%
    \begin{tabular}{l|r|rr|rr|rr}
    \toprule
    \midrule    
    Ratio & Criterion & WikiText2$\downarrow$ & $\Delta \uparrow$  & PTB$\downarrow$ & $\Delta \uparrow$ & Avg. $\uparrow$ & $\Delta \uparrow$\\ 
    \midrule 
    \multirow{2}{*}{20\%} & 
    $| W \hat{\nabla} \mathcal{L}(W) |$ & 20.40 & \multirow{2}{*}{2.08} & 34.17 & \multirow{2}{*}{1.63} & 67.24 & \multirow{2}{*}{0.33}\\
    & $| W \hat{\nabla} \mathcal{L}(W) X |$ & \textbf{18.32} & & \textbf{32.54} & & \textbf{67.57} &\\
    \midrule 
    \multirow{2}{*}{30\%} & 
    $| W \hat{\nabla}\mathcal{L}(W) |$ & 25.09 & \multirow{2}{*}{0.81} & 40.89 & \multirow{2}{*}{1.87} & 61.97 & \multirow{2}{*}{2.02}\\
    & $| W \hat{\nabla}\mathcal{L}(W) X|$ & \textbf{24.28} & & \textbf{39.02} & & \textbf{63.99}\\
    \midrule 
    \multirow{2}{*}{40\%} & 
    $|W \hat{\nabla} \mathcal{L}(W)|$ & 35.25 & \multirow{2}{*}{3.47} &52.30  & \multirow{2}{*}{3.07} & 58.95 & \multirow{2}{*}{2.07}\\
    & $|W \hat{\nabla} \mathcal{L}(W) X|$ & \textbf{31.78} & & \textbf{49.23} & & \textbf{61.02}\\
    \midrule 
    \multirow{2}{*}{50\%} & 
    $|W \hat{\nabla}\mathcal{L}(W)|$ & 53.23 & \multirow{2}{*}{8.54} & 75.50 & \multirow{2}{*}{5.67} & 54.73 & \multirow{2}{*}{2.20}\\
    & $|W \hat{\nabla}\mathcal{L}(W) X|$ & \textbf{44.69} & & \textbf{69.83} & & \textbf{56.93}\\
    \midrule    
    \bottomrule
    \end{tabular}
    }
   \vspace{-1mm}
   \caption{Zero-shot performance of the pruned LLaMA-7B models achieved by using ZO gradient-based pruning criterion with/without activations. The columns have the same meaning as Table~\ref{tab:taylor-with-vs-without-activations}.}
   \label{tab:fwdgrad-with-vs-without-activations}
   \vspace{-2mm}
\end{table}

\vspace{0.2cm}
\noindent \textbf{Efficacy of activations on BLOOM-7B and OPT-6.7B}
Table~\ref{tab:taylor-with-vs-without-activations-bloom-opt}, Table~\ref{tab:fwdgrad-with-vs-without-activations-bloom-without-tune} and Table~\ref{tab:fwdgrad-with-activations-OPT}
show the zero-shot results of the pruned models with/without fine-tuning to discover the effectiveness of activations for guiding pruning LLMs. From the results in Table~\ref{tab:taylor-with-vs-without-activations-bloom-opt}, it is evident that incorporating activations $| W\nabla \mathcal{L}(W) X |$ results in an overall enhancement of performance compared to the standard pruning criterion $| W\nabla \mathcal{L}(W) |$. For example, at a 10\% pruning ratio, the model pruned with $| W\nabla \mathcal{L}(W) X |$ achieves a perplexity of 35.53 on the WikiText2 dataset for pruning BLOOM-7B. This result surpasses the 38.12 perplexity achieved by the model compressed by using $| W\nabla \mathcal{L}(W) |$. For OPT-6.7B, the performance enhancement provided by activation is also noticeable. For instance, with 30\% parameters pruned, the model pruned with $ | W\nabla \mathcal{L}(W) X |$ achieves a perplexity of 64.58 on PTB. Compared to using $| W\nabla \mathcal{L}(W) |$, the perplexity result decreased by 0.51. 

\begin{table}[t]
    \centering
    \small
    \scalebox{0.89}{%
    \begin{tabular}{l|r|rr|rr}
    \toprule
    \midrule    
    Prune Ratio & Model/Criterion & WikiText2$\downarrow$ & $\Delta \uparrow$  & PTB$\downarrow$ & $\Delta \uparrow$ \\ 
    \midrule
    0\% & BLOOM-7B & 26.58 & - & 50.55 & -\\
    \midrule
    10\% & 
    $| W \nabla \mathcal{L}(W) |$ & 38.12 & \multirow{2}{*}{2.59} & 86.23 & \multirow{2}{*}{8.63}\\
    w/ tune & $| W \nabla \mathcal{L}(W) X |$ & \textbf{35.53} & & \textbf{77.60} & \\
    \midrule 
    30\% & 
    $| W \nabla\mathcal{L}(W) |$ & \textbf{58.11} & \multirow{2}{*}{-0.24} & 147.52 & \multirow{2}{*}{9.72}\\
    w/ tune & $| W \nabla \mathcal{L}(W) X|$ & 58.35 & & \textbf{137.80} & \\
    \midrule
    0\% & OPT-6.7B & 26.45 & - & 32.03 & -\\
    \midrule
    10\% & 
    $|W \nabla \mathcal{L}(W)|$ & \textbf{27.89} & \multirow{2}{*}{-0.38} & 39.33  & \multirow{2}{*}{0.39}\\
    w/ tune& $|W \nabla \mathcal{L}(W) X|$ & 28.27 & & \textbf{38.94} &\\
    \midrule 
    30\% & 
    $|W \nabla \mathcal{L}(W)|$ & 42.94 & \multirow{2}{*}{0.75} & 65.09 & \multirow{2}{*}{0.51}\\
    w/ tune & $|W \nabla \mathcal{L}(W) X|$ & \textbf{42.19} & & \textbf{64.58} &\\
    \midrule
    \bottomrule
    \end{tabular}
    }
   \vspace{-1mm}    
   \caption{Zero-shot perplexity of the pruned BLOOM-7B and OPT-6.7B models achieved by using backpropagation gradient-based pruning criterion with/without activations. The columns have the same meaning as Table~\ref{tab:gradient-based-gradient-free-bloom-opt}.}
   \vspace{-2mm}   
    \label{tab:taylor-with-vs-without-activations-bloom-opt}
\end{table}

In Table~\ref{tab:taylor-with-vs-without-activations-bloom-opt}, gradients are computed by backpropagation. In contrast, the results presented in Table~\ref{tab:fwdgrad-with-vs-without-activations-bloom-without-tune} and \ref{tab:fwdgrad-with-activations-OPT} highlight the effectiveness of incorporating activations in estimated gradient-based pruning criteria when fine-tuning is applied or not. For instance, when fine-tuning is not applied, the model pruned with 30\% of BLOOM-7B's parameters using $|W \hat{\nabla} \mathcal{L}(W) X|$ achieves a perplexity of 91.43 on WikiText2, reflecting a 14.64 improvement over the 106.07 perplexity by the model using $|W \hat{\nabla} \mathcal{L}(W)|$. In contrast, after undergoing fine-tuning, although the increase in performance becomes less pronounced as shown in Table~\ref{tab:fwdgrad-with-activations-OPT}, it is still evident that activations play a significant role in enhancing performance.

\begin{table}[t]
    \centering
    \small
    \scalebox{0.84}{%
    \begin{tabular}{l|r|rr|rr}
    \toprule
    \midrule    
    Prune Ratio & Model/Criterion & WikiText2$\downarrow$ & $\Delta \uparrow$  & PTB$\downarrow$ & $\Delta \uparrow$ \\ 
    \midrule
    0\% & BLOOM-7B & 26.58 & - & 50.55 & -\\
    \midrule
    10\% & 
    $| W \hat{\nabla} \mathcal{L}(W) |$ & 78.21 & \multirow{2}{*}{4.74} & 231.67 & \multirow{2}{*}{27.22}\\
    w/o tune & $| W \hat{\nabla} \mathcal{L}(W) X |$ & \textbf{73.47} & & \textbf{204.45} & \\
    \midrule 
    30\% & 
    $| W \hat{\nabla} \mathcal{L}(W) |$ & 106.07 & \multirow{2}{*}{14.64} & 239.96.52 & \multirow{2}{*}{31.48}\\
    w/o tune & $| W \hat{\nabla} \mathcal{L}(W) X|$ & \textbf{91.43} & & \textbf{208.48} & \\
    \midrule        
    \bottomrule
    \end{tabular}
   }
   \vspace{-1mm}   
   \caption{Zero-shot perplexity of the pruned BLOOM-7B achieved by using ZO gradient-based pruning criterion with/without activations. The columns have the same meaning as Table~\ref{tab:gradient-based-gradient-free-bloom-opt}.}
   \vspace{-2mm}   
   \label{tab:fwdgrad-with-vs-without-activations-bloom-without-tune}
\end{table}

\begin{table}[t]
    \centering
    \small
    \scalebox{0.89}{%
    \begin{tabular}{l|r|rr|rr}
    \toprule
    \midrule
    Prune Ratio & Model/Criterion & WikiText2$\downarrow$ & $\Delta \uparrow$  & PTB$\downarrow$ & $\Delta \uparrow$ \\ 
    \midrule
    0\% & OPT-6.7B & 26.45 & - & 32.03 & -\\
    \midrule
    10\% & 
    $| W \hat{\nabla} \mathcal{L}(W) |$ & 30.51 & \multirow{2}{*}{0.36} & 39.17 & \multirow{2}{*}{0.53}\\
    w/ tune & $| W \hat{\nabla} \mathcal{L}(W) X |$ & \textbf{30.15} & & \textbf{38.64} & \\
    \midrule 
    30\% & 
    $| W \hat{\nabla} \mathcal{L}(W) |$ & 43.87 & \multirow{2}{*}{2.98} &58.12 & \multirow{2}{*}{0.68}\\
    w/ tune & $| W \hat{\nabla} \mathcal{L}(W) X|$ & \textbf{40.89} & & \textbf{57.44} & \\
    \midrule    
    \bottomrule
    \end{tabular}
    }
   \vspace{-1mm}    
   \caption{Zero-shot perplexity of the pruned OPT-6.7B achieved by using ZO gradient-based pruning criterion with/without activations. The columns have the same meaning as Table~\ref{tab:gradient-based-gradient-free-bloom-opt}.}
   \vspace{-2mm}   
  \label{tab:fwdgrad-with-activations-OPT}
\end{table}

\begin{figure}[t]
  \centering
  \vspace{-3mm}    
  \subfloat[MINI-LLM (ours)]{
  \begin{minipage}{4cm}
  \includegraphics[width=4cm,height=3cm]{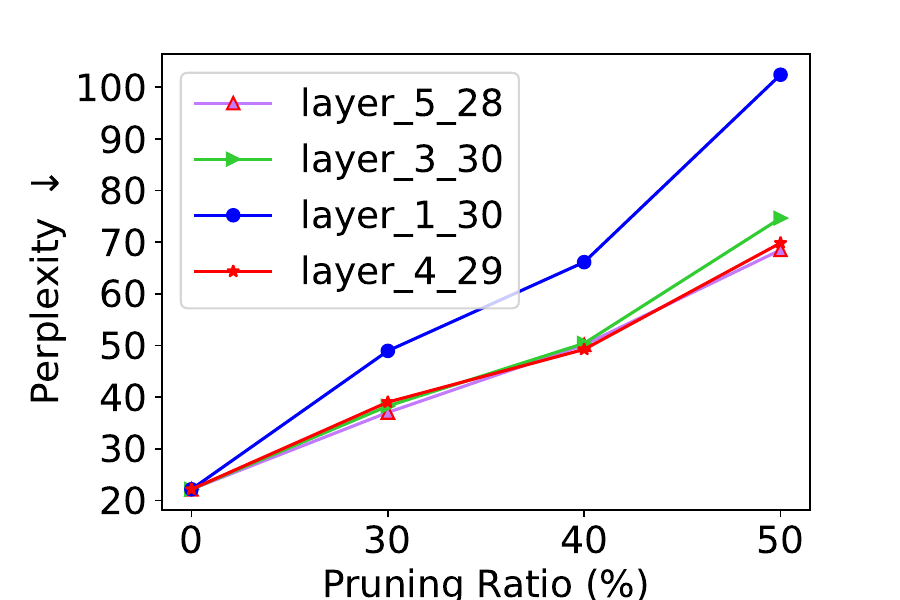}
     \label{Fig:llama-layer-sensitivity-fwdgrad-ptb}
  \end{minipage}
  }
  \subfloat[LLM-Pruner]{  
  \begin{minipage}{4cm}
  \includegraphics[width=4cm,height=3cm]{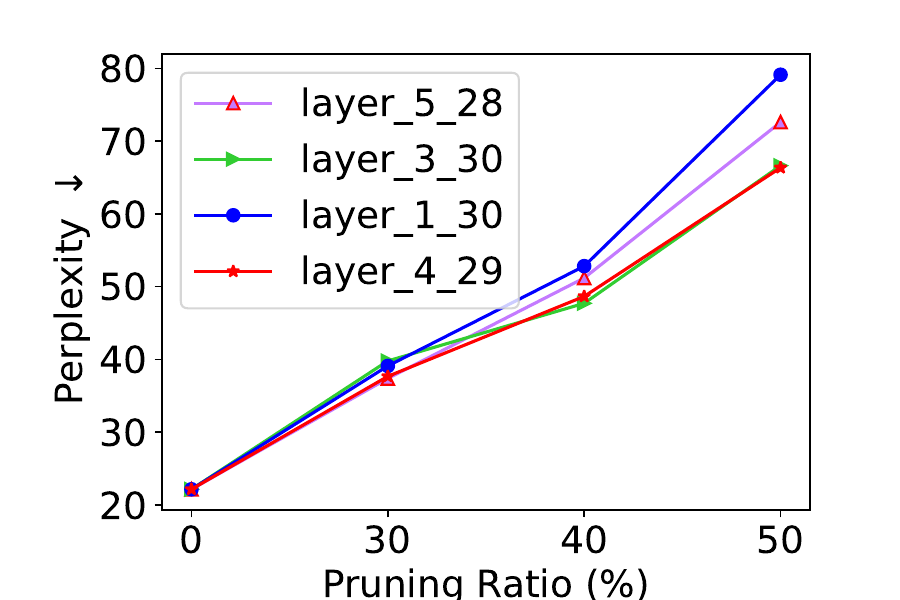}
     \label{Fig:llama-layer-sensitivity-taylor-ptb}
  \end{minipage} 
  } 
 \vspace{-0.1cm}  
 \caption{The zero-shot perplexity of the pruned models achieved by enabling different ranges of layers involved in pruning LLaMA-7B on the PTB dataset. Layer-1-30 has the worst performance, while layer-3-30 and layer-4-29 have comparably better performance for both pruning methods. In contrast, the models derived from layer-5-28 pruning exhibit varying responses to different pruning methods.}
 \vspace{-0.2cm} 
 \label{Fig:llama-layer-sensitivity}
\end{figure}

\vspace{0.2cm}
\noindent \textbf{Layer Sensitivity for Pruning}
Based on the findings in~[\cite{ma2023llmpruner}] that the first and last layers significantly affect the model's performance, we investigate the impact of involving different ranges of layers in the pruning process on LLaMA-7B's performance. It includes analyzing the performance of models with pruning applied from the 1st to the 30th layer (represented by layer-1-30 in Figure~\ref{Fig:llama-layer-sensitivity}), from the 3rd to the 30th layer (layer-3-30), from the 4th to the 29th layer (layer-4-29), and from the 5th to the 28th layer (layer-5-28)~\footnote{LLaMA-7B has 32 layers, from 0th to the 31st layer.}. From the results in Figure~\ref{Fig:llama-layer-sensitivity}, it is evident that layer-1-30 has the worst performance, while layer-3-30 and layer-4-29 have comparably better performance for both pruning methods. In contrast, the models derived from layer-5-28 pruning exhibit varying responses to different pruning methods. For instance, for MINI-LLM, their performance is similar to that of the layer-4-29 models, whereas, for LLM-Pruner~[\cite{ma2023llmpruner}], their performance is somewhat inferior compared to both layer-4-29 and layer-3-30 models. Since the layer-4-29 pruning demonstrates consistent performance across various pruning methods, we conduct pruning for layer-4-29 in all LLaMA-7B experiments.

\begin{figure}[t]
  \centering
  \vspace{-3mm}    
  \subfloat[WikiText2]{
  \begin{minipage}{4cm}
  \includegraphics[width=4cm,height=3cm]{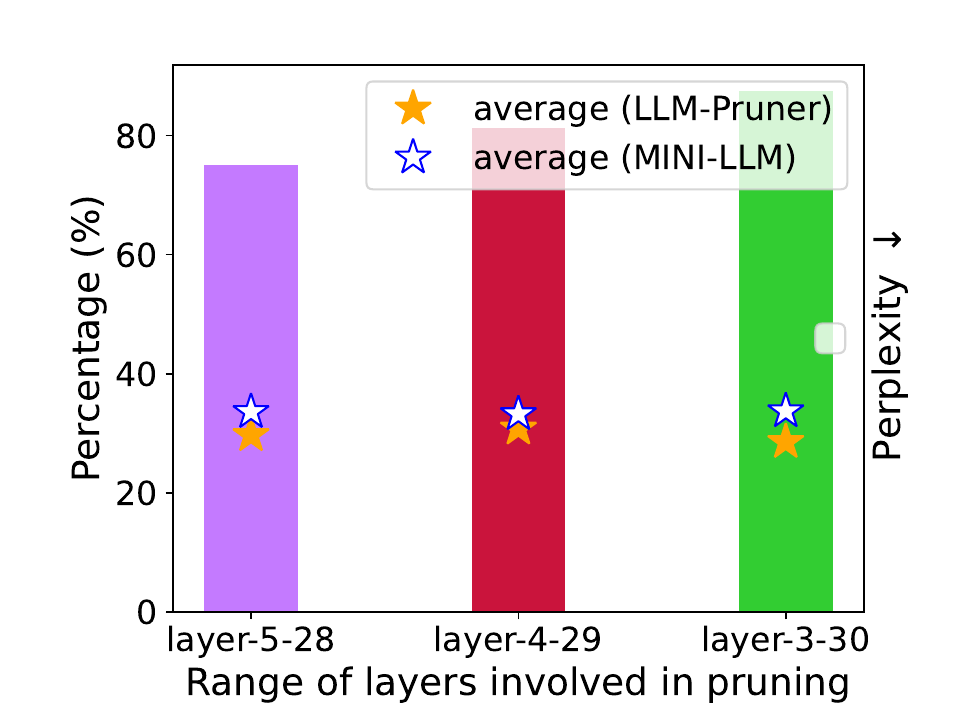}
     \label{Fig:llama-layer-sensitivity-average-wikitext2}
  \end{minipage}
  }
  \subfloat[PTB]{  
  \begin{minipage}{4cm}
  \includegraphics[width=4cm,height=3cm]{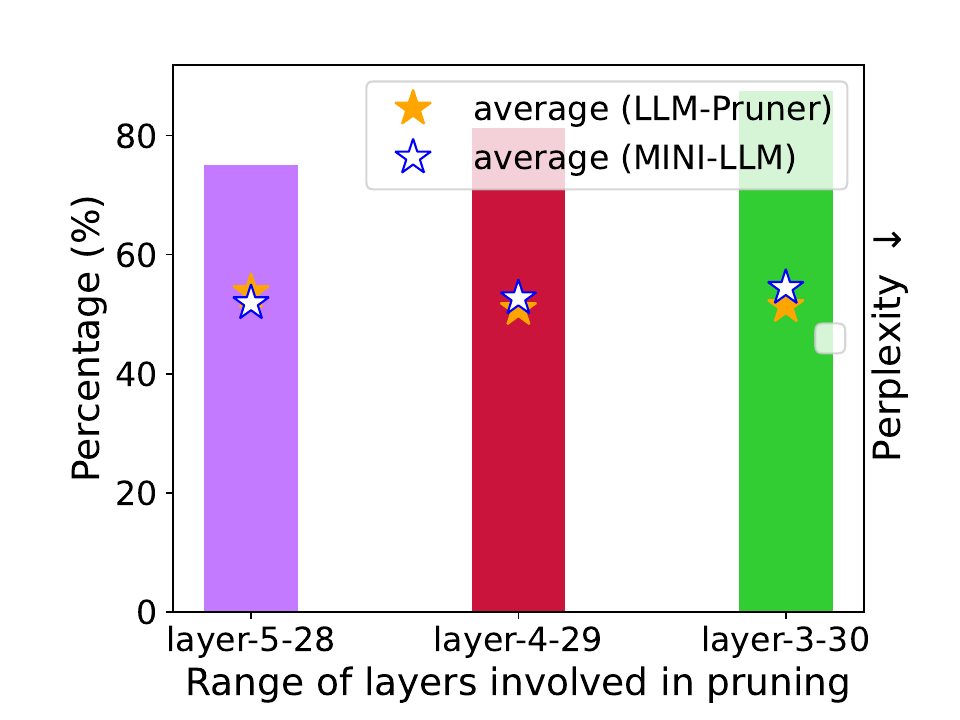}
     \label{Fig:llama-layer-sensitivity-average-ptb}
  \end{minipage} 
  }  
 \vspace{-0.1cm}  
 \caption{The percentage of different ranges of layers involved in pruning LLaMA-7B and the average perplexity. MINI-LLM exhibits performance close to, even surpasses LLM-Pruner.}
 \vspace{-0.2cm}
 \label{Fig:llama-layer-sensitivity-average}
\end{figure}

Considering the inferior performance of layer-1-31, in Figure~\ref{Fig:llama-layer-sensitivity-average}, we only display the proportions of layers involved in pruning for layer-3-30, layer-4-29, and layer-5-28. In addition, we present the average perplexity of LLM-Pruner and MINI-LLM for the three ranges of layers, marked with the orange or the blue five-pointed stars in Figure~\ref{Fig:llama-layer-sensitivity-average}, which is averaged over three pruning ratios: 30\%, 40\%, and 50 \% on WikiText2 and PTB. The average perplexity results in Figure~\ref{Fig:llama-layer-sensitivity-average} illustrate that our MINI-LLM exhibits performance close to, even surpasses at times, that of LLM-Pruner. For instance, for layer-5-28, MINI-LLM achieves an average perplexity of 51.87 on PTB, outperforming LLM-Pruner's 53.73. These results demonstrate that MINI-LLM is a memory-efficient and effective method for gradient-based pruning.

\subsection{Generations From Pruned Model}
Table~\ref{tab:llama-7b-generation} shows the generation examples of the original and the pruned LLaMA-7B models achieved by MINI-LLM. The five experimental instructions encompass math, common sense, translation, and writing tasks. From the responses presented in Table~\ref{tab:llama-7b-generation}, it is evident that when pruning 20\% of the parameters, the pruned model maintains high performance in these tasks.

\begin{table*}[t]
    \centering
    \small
    \renewcommand{\arraystretch}{1.5}     
    \scalebox{0.80}{%
    \begin{tabular}{p{2cm}|p{15cm}}
        \toprule
        Model  & Generated Sentences\\
        \midrule
        \multirow{17}{2cm}{LLaMA-7B (ratio=0\%)} & \underline{Instruction:} What are Fibonacci numbers? \underline{Response:} Fibonacci numbers is a sequence of integer numbers, generated by a simple formula which has the form of sum of two numbers. This sequence has a few very interesting properties which can be summarized as follows: Every Fibonacci number is the sum of two preceding Fibonacci numbers. The formula is Fn= Fn-1 + Fn-2. Start with 1,2,3,5,8,13,21,34,55,89,… \\
        \cline{2-2} 
        & \underline{Instruction:} What is the difference between a cat and a tiger? \underline{Response:} A tiger is a member of the cat family, but is a big, powerful animal with striped fur. It is bigger than a normal housecat. The tiger is native to India and southwestern Asia. In contrast, a cat is smaller than the tiger. There are three main types of cats. Most cats are of the domestic.  \\
        \cline{2-2}
        & \underline{Instruction:} AI can create a logo in seconds \underline{Response:} When you think about a logo, the first thought may be to sit down and draw something, or even create a vector file. But the truth is that it is possible to do this easily and quickly by using AI. So, the question is, why not use it? For example, I was recently working with the logo of an Italian company and I only needed to sit down at the computer for 5 minutes to create the basic elements that I would then change to meet the needs of the brand. In just 2 minutes you can create a 3D logo. \\
        \cline{2-2}
        & \underline{Instruction:} Compose a poem about summer. \underline{Response:} Sandy toes, Cooling ocean, Water lilies, the warm sun, Making the clouds, dissolve, and the birds sing. Their songs lull me to sleep, With soft breeze. Summer is, the best, because there is no one to bother, in the summer days. \\
        \cline{2-2}
        & \underline{Instruction:} Translate the English sentence "large language models" into Spanish. \underline{Response:} The answer for the English sentence is "grandes modelos lenguas" in Spanish. \\
        \midrule
        \multirow{20}{2cm}{LLaMA-7B (ratio=20\%)} & \underline{Instruction:} What are Fibonacci numbers? \underline{Response:} In mathematics and physics, the Fibonacci numbers are the sequence of integers starting from the 2nd number as the sum of the previous two numbers in the sequence: 1, 1, 2, 3, 5, 8, 13, 21, 34, 55,… \\
        \cline{2-2} 
        & \underline{Instruction:} What is the difference between a cat and a tiger? \underline{Response:} The cat and the tiger are both Felines, which means both species belong to the same Biological Taxonomic Classification family. However, the difference between a cat and a tiger lies primarily in the physical differences between the two species. A cat is generally smaller in size, with an average weight of 5 - 10 pounds (2.7 - 4.4 kilograms). Cats are more domesticated, often being kept as pets. They are known for being more social, and their distinctive meowing, hisses, and growls are used to communicate with other felines. On the other hand, tigers are bigger in size, with an average weight of around 125 pounds (69.5 kilograms). They are also more wild, with little domestication.  \\
        \cline{2-2}
        & \underline{Instruction:} AI can create a logo in seconds \underline{Response:} AI can create a logo in seconds, a visual designer can bring your brand idea to life. When it comes to creating a brand identity, our first step is to conduct a thorough market analysis and develop a brand position. From there we can create a creative brand identity that is in line with the core values of your business. Our creative design team is trained in building brands that are aesthetically pleasing, easy to recognize and communicate a clear message to your target audience. The design can incorporate many elements including illustrations, photography, infographics, videos, typography and layout. \\
        \cline{2-2}
        & \underline{Instruction:} Compose a poem about summer. \underline{Response:} A summer sun is all aglow, Reflecting in the calm, clear waters below, Tossed by gentle breeze, Warm and golden, Far beyond sight. A perfect sight for sight, In an endless horizon, Reflecting in the calm, Clear waters below. A summer sun, That never sets, Far beyond sight. \\
        \cline{2-2}
        & \underline{Instruction:} Translate the English sentence "large language models" into Spanish. \underline{Response:} Here's a suggested sentence in Spanish: "Modelo grande de lenguas." \\
        \bottomrule
    \end{tabular}
    }
    \caption{Generated Examples from the original and pruned LLaMA-7B.}
    \label{tab:llama-7b-generation}
\end{table*}

\section{Conclusion}
In this paper, we presented MINI-LLM, an one-shot structured pruning approach designed to address the high GPU memory demands of computing backpropagation-based gradients of pre-trained LLMs. First, we proposed a novel criterion called the Feature Map Sensitivity (FMS) score which integrates magnitude, activation, and gradient information to guide the pruning process effectively. By employing estimated gradients based on forward passes, MINI-LLM not only reduces the GPU memory requirement for gradient-guided pruning but also achieves superior performance compared to existing gradient-free methods. Our extensive experiments on three LLMs: LLaMA, BLOOM, and OPT, across various downstream tasks demonstrate MINI-LLM's effectiveness and efficiency in GPU memory usage. In the future, our objective is to further enhance the pruning results of MINI-LLM at higher compression rates. 

\bibliographystyle{named}
\bibliography{ref}

\begin{thebibliography}{}

\bibitem[\protect\citeauthoryear{Bisk \bgroup \em et al.\egroup }{2020}]{bisk2020piqa}
Yonatan Bisk, Rowan Zellers, Ronan~Le Bras, Jianfeng Gao, and Yejin Choi.
\newblock {PIQA}: Reasoning about physical commonsense in natural language.
\newblock In {\em AAAI}, 2020.

\bibitem[\protect\citeauthoryear{Brown \bgroup \em et al.\egroup }{2020}]{brown2020language}
Tom Brown, Benjamin Mann, Nick Ryder, et~al.
\newblock Language models are few-shot learners.
\newblock {\em Advances in neural information processing systems}, 33:1877--1901, 2020.

\bibitem[\protect\citeauthoryear{Chavan \bgroup \em et al.\egroup }{2023}]{chavan2023oneforall}
Arnav Chavan, Zhuang Liu, Deepak Gupta, Eric Xing, and Zhiqiang Shen.
\newblock {One-for-All}: Generalized lora for parameter-efficient fine-tuning.
\newblock {\em arXiv preprint arXiv:2306.07967}, 2023.

\bibitem[\protect\citeauthoryear{Cheng \bgroup \em et al.\egroup }{2023}]{cheng2023survey}
Hongrong Cheng, Miao Zhang, and Javen~Qinfeng Shi.
\newblock A survey on deep neural network pruning-taxonomy, comparison, analysis, and recommendations.
\newblock {\em arXiv preprint arXiv:2308.06767}, 2023.

\bibitem[\protect\citeauthoryear{Clark \bgroup \em et al.\egroup }{2019}]{clark2019boolq}
Christopher Clark, Kenton Lee, Ming-Wei Chang, Tom Kwiatkowski, Michael Collins, and Kristina Toutanova.
\newblock {BoolQ}: Exploring the surprising difficulty of natural yes/no questions.
\newblock In {\em NAACL}, 2019.

\bibitem[\protect\citeauthoryear{Dettmers \bgroup \em et al.\egroup }{2022}]{dettmers2022llmint8}
Tim Dettmers, Mike Lewis, Younes Belkada, and Luke Zettlemoyer.
\newblock {LLM.int8()}: 8-bit matrix multiplication for transformers at scale.
\newblock In {\em NeurIPS}, 2022.

\bibitem[\protect\citeauthoryear{Dettmers \bgroup \em et al.\egroup }{2023}]{dettmers2023qlora}
Tim Dettmers, Artidoro Pagnoni, Ari Holtzman, and Luke Zettlemoyer.
\newblock {QLoRA}: Efficient finetuning of quantized {LLMs}.
\newblock {\em arXiv preprint arXiv:2305.14314}, 2023.

\bibitem[\protect\citeauthoryear{Frantar and Alistarh}{2023}]{frantar2023sparsegpt}
Elias Frantar and Dan Alistarh.
\newblock {SparseGPT}: Massive language models can be accurately pruned in one-shot.
\newblock {\em arXiv preprint arXiv:2301.00774}, 2023.

\bibitem[\protect\citeauthoryear{Frantar \bgroup \em et al.\egroup }{2022}]{frantar2022optimal}
Elias Frantar, Sidak~Pal Singh, and Dan Alistarh.
\newblock Optimal brain compression: A framework for accurate post-training quantization and pruning.
\newblock In {\em NeurIPS}, 2022.

\bibitem[\protect\citeauthoryear{Fu \bgroup \em et al.\egroup }{2022}]{fu2022depthshrinker}
Yonggan Fu, Haichuan Yang, Jiayi Yuan, Meng Li, Cheng Wan, Raghuraman Krishnamoorthi, Vikas Chandra, and Yingyan Lin.
\newblock Depthshrinker: a new compression paradigm towards boosting real-hardware efficiency of compact neural networks.
\newblock In {\em ICML}, 2022.

\bibitem[\protect\citeauthoryear{Gasnikov \bgroup \em et al.\egroup }{2022}]{gasnikov2022randomized}
Alexander Gasnikov, Darina Dvinskikh, Pavel Dvurechensky, Eduard Gorbunov, Aleksander Beznosikov, and Alexander Lobanovu.
\newblock Randomized gradient-free methods in convex optimization.
\newblock {\em arXiv preprint arXiv:2211.13566}, 2022.

\bibitem[\protect\citeauthoryear{He \bgroup \em et al.\egroup }{2023}]{he2023sensitivity}
Haoyu He, Jianfei Cai, Jing Zhang, Dacheng Tao, and Bohan Zhuang.
\newblock Sensitivity-aware visual parameter-efficient tuning.
\newblock In {\em ICCV}, 2023.

\bibitem[\protect\citeauthoryear{Hendrycks \bgroup \em et al.\egroup }{2021}]{hendrycks2021measuring}
Dan Hendrycks, Collin Burns, Steven Basart, Andy Zou, Mantas Mazeika, Dawn Song, and Jacob Steinhardt.
\newblock Measuring massive multitask language understanding.
\newblock In {\em International Conference on Learning Representations (ICLR)}, 2021.

\bibitem[\protect\citeauthoryear{Hu \bgroup \em et al.\egroup }{2022}]{hu2022lora}
Edward~J. Hu, Yelong Shen, Phillip Wallis, Zeyuan Allen-Zhu, Yuanzhi Li, Shean Wang, and Weizhu Chen.
\newblock {LoRA}: Low-rank adaptation of large language models.
\newblock In {\em ICLR poster}, 2022.

\bibitem[\protect\citeauthoryear{Jia \bgroup \em et al.\egroup }{2022}]{jia2022visual}
Menglin Jia, Luming Tang, Bor-Chun Chen, Claire Cardie, Serge Belongie, Bharath Hariharan, and Ser-Nam Lim.
\newblock Visual prompt tuning.
\newblock In {\em ECCV}, 2022.

\bibitem[\protect\citeauthoryear{Kiefer and Wolfowitz.}{1952}]{kiefer1952stochastic}
J.~Kiefer and J.~Wolfowitz.
\newblock Stochastic estimation of the maximum of a regression function.
\newblock {\em Ann. Math. Statist}, 23:462--466, 1952.

\bibitem[\protect\citeauthoryear{Kingma and Ba}{2015}]{kingma2015adam}
Diederik~P. Kingma and Jimmy Ba.
\newblock {Adam}: A method for stochastic optimization.
\newblock In {\em ICLR}, 2015.

\bibitem[\protect\citeauthoryear{Kovaleva \bgroup \em et al.\egroup }{2021}]{kovaleva2021bertbusters}
Olga Kovaleva, Saurabh Kulshreshtha, Anna Rogers, and Anna Rumshisky.
\newblock {BERT Busters}: Outlier dimensions that disrupt transformers.
\newblock In {\em ACL}, 2021.

\bibitem[\protect\citeauthoryear{Kurtic \bgroup \em et al.\egroup }{2022}]{kurtic2022optimal}
Eldar Kurtic, Daniel Campos, Tuan Nguyen, Elias Frantar, Mark Kurtz, Benjamin Fineran, Michael Goin, and Dan Alistarh.
\newblock The optimal {BERT} surgeon: Scalable and accurate second-order pruning for large language models.
\newblock In {\em EMNLP}, 2022.

\bibitem[\protect\citeauthoryear{Kurtic \bgroup \em et al.\egroup }{2023}]{kurtic2023ziplm}
Eldar Kurtic, Elias Frantar, and Dan Alistarh.
\newblock {ZipLM}: Inference-aware structured pruning of language models.
\newblock In {\em NeurIPS}, 2023.

\bibitem[\protect\citeauthoryear{Kwon \bgroup \em et al.\egroup }{2022}]{kwon2022fast}
Woosuk Kwon, Sehoon Kim, Michael~W. Mahoney, Joseph Hassoun, Kurt Keutzer, and Amir Gholami.
\newblock A fast post-training pruning framework for transformers.
\newblock In {\em NeurIPS}, 2022.

\bibitem[\protect\citeauthoryear{LeCun \bgroup \em et al.\egroup }{1989}]{lecun1989optimal}
Yann LeCun, John Denker, and Sara Solla.
\newblock Optimal brain damage.
\newblock In {\em NIPS}, pages 598--605, 1989.

\bibitem[\protect\citeauthoryear{Lee \bgroup \em et al.\egroup }{2019}]{lee2019snip}
Namhoon Lee, Thalaiyasingam Ajanthan, and Philip H.~S. Torr.
\newblock {SNIP}: Single-shot network pruning based on connection sensitivity.
\newblock In {\em ICLR}, 2019.

\bibitem[\protect\citeauthoryear{Lester \bgroup \em et al.\egroup }{2021}]{lester2021power}
Brian Lester, Rami Al-Rfou, and Noah Constant.
\newblock The power of scale for parameter-efficient prompt tuning.
\newblock In {\em EMNLP}, 2021.

\bibitem[\protect\citeauthoryear{Li and Liang}{2021}]{li2021prefixtuning}
Xiang~Lisa Li and Percy Liang.
\newblock {Prefix-Tuning}: Optimizing continuous prompts for generation.
\newblock In {\em IJCNLP}, 2021.

\bibitem[\protect\citeauthoryear{Li \bgroup \em et al.\egroup }{2020}]{li2020group}
Yawei Li, Shuhang Gu, Christoph Mayer, Luc~Van Gool, and Radu Timofte.
\newblock Group sparsity: The hinge between filter pruning and decomposition for network compression.
\newblock In {\em CVPR}, 2020.

\bibitem[\protect\citeauthoryear{Li \bgroup \em et al.\egroup }{2022a}]{li2022simultaneous}
Shiru Li, Yong Xia, and Zi~Xu.
\newblock Simultaneous perturbation stochastic approximation: towards one-measurement per iteration.
\newblock {\em arXiv preprint arXiv:2203.03075}, 2022.

\bibitem[\protect\citeauthoryear{Li \bgroup \em et al.\egroup }{2022b}]{li2022parameter}
Yuchao Li, Fuli Luo, Chuanqi Tan, Mengdi Wang, Songfang Huang, Shen Li, and Junjie Bai.
\newblock Parameter-efficient sparsity for large language models fine-tuning.
\newblock In {\em IJCAI}, 2022.

\bibitem[\protect\citeauthoryear{Liu \bgroup \em et al.\egroup }{2021}]{liu2021group}
Liyang Liu, Shilong Zhang, Zhanghui Kuang, Aojun Zhou, Jing-Hao Xue, Xinjiang Wang, Yimin Chen, Wenming Yang, Qingmin Liao, and Wayne Zhang.
\newblock Group fisher pruning for practical network compression.
\newblock In {\em ICML}, 2021.

\bibitem[\protect\citeauthoryear{Ma \bgroup \em et al.\egroup }{2023}]{ma2023llmpruner}
Xinyin Ma, Gongfan Fang, and Xinchao Wang.
\newblock {LLM-Pruner}: On the structural pruning of large language models.
\newblock In {\em NeurIPS}, 2023.

\bibitem[\protect\citeauthoryear{Malladi \bgroup \em et al.\egroup }{2023}]{malladi2023finetuning}
Sadhika Malladi, Tianyu Gao, Eshaan Nichani, Alex Damian, Jason~D. Lee, Danqi Chen, and Sanjeev Arora.
\newblock Fine-tuning language models with just forward passes.
\newblock In {\em NeurIPS}, 2023.

\bibitem[\protect\citeauthoryear{Marcus \bgroup \em et al.\egroup }{1993}]{marcus1993building}
Mitchell Marcus, Beatrice Santorini, and Mary~Ann Marcinkiewicz.
\newblock Building a large annotated corpus of {English}: The penn treebank.
\newblock {\em Computational Linguistics}, 19:313--330, 1993.

\bibitem[\protect\citeauthoryear{Merity \bgroup \em et al.\egroup }{2016}]{merity2016pointer}
Stephen Merity, Caiming Xiong, James Bradbury, and Richard Socher.
\newblock Pointer sentinel mixture models.
\newblock {\em arXiv preprint arXiv:1609.07843}, 2016.

\bibitem[\protect\citeauthoryear{Molchanov \bgroup \em et al.\egroup }{2017}]{molchanov2017pruning}
Pavlo Molchanov, Stephen Tyree, Tero Karras, Timo Aila, and Jan Kautz.
\newblock Pruning convolutional neural networks for resource-efficient inference.
\newblock In {\em ICLR}, 2017.

\bibitem[\protect\citeauthoryear{Molchanov \bgroup \em et al.\egroup }{2019}]{molchanov2019importance}
Pavlo Molchanov, Arun Mallya, Stephen Tyree, Iuri Frosio, and Jan Kautz.
\newblock Importance estimation for neural network pruning.
\newblock In {\em CVPR}, 2019.

\bibitem[\protect\citeauthoryear{Nova \bgroup \em et al.\egroup }{2023}]{nova2023gradient}
Azade Nova, Hanjun Dai, and Dale Schuurmans.
\newblock Gradient-free structured pruning with unlabeled data.
\newblock In {\em ICML}, 2023.

\bibitem[\protect\citeauthoryear{OpenAI}{2023}]{open2023gpt4}
OpenAI.
\newblock {GPT-4} technical report.
\newblock {\em arXiv preprint arXiv:2303.08774}, 2023.

\bibitem[\protect\citeauthoryear{Sakaguchi \bgroup \em et al.\egroup }{2021}]{sakaguchi2021winogrande}
Keisuke Sakaguchi, Ronan~Le Bras, Chandra Bhagavatula, and Yejin Choi.
\newblock {WinoGrande}:an adversarial winograd schema challenge at scale.
\newblock {\em Communications of the ACM}, 64:99--106, 2021.

\bibitem[\protect\citeauthoryear{Sanh \bgroup \em et al.\egroup }{2020}]{sanh2020movement}
Victor Sanh, Thomas Wolf, and Alexander~M. Rush.
\newblock Movement pruning: Adaptive sparsity by fine-tuning.
\newblock In {\em NeurIPS}, 2020.

\bibitem[\protect\citeauthoryear{Spall}{1992}]{spall1992multivariate}
James~C. Spall.
\newblock Multivariate stochastic approximation using a simultaneous perturbation gradient approximation.
\newblock {\em IEEE Transactions on Automatic Control}, 37:332--341, 1992.

\bibitem[\protect\citeauthoryear{Spall}{1997}]{spall1997onemeasurement}
James~C. Spall.
\newblock A one-measurement form of simultaneous perturbation stochastic approximation.
\newblock {\em Automatics}, 33:109--112, 1997.

\bibitem[\protect\citeauthoryear{Sun \bgroup \em et al.\egroup }{2024}]{sun2024simple}
Mingjie Sun, Zhuang Liu, Anna Bair, and J.~Zico Kolter.
\newblock A simple and effective pruning approach for large language models.
\newblock In {\em Proceedings of International Conference on Learning Representations (ICLR) poster}, 2024.

\bibitem[\protect\citeauthoryear{Taori \bgroup \em et al.\egroup }{2023}]{taori2023stanford}
Rohan Taori, Ishaan Gulrajani, Tianyi Zhang, Yann Dubois, Xuechen Li, Carlos Guestrin, Percy Liang, and Tatsunori~B. Hashimoto.
\newblock Stanford alpaca: An instruction-following llama model.
\newblock \url{https://github.com/tatsu-lab/stanford_alpaca}, 2023.

\bibitem[\protect\citeauthoryear{Touvron \bgroup \em et al.\egroup }{2023}]{touvron2023llama}
Hugo Touvron, Thibaut Lavril, Gautier Izacard, et~al.
\newblock {LLaMA}: Open and efficient foundation language models.
\newblock {\em arXiv preprint arXiv:2302.13971}, 2023.

\bibitem[\protect\citeauthoryear{Wang \bgroup \em et al.\egroup }{2020a}]{wang2020picking}
Chaoqi Wang, Guodong Zhang, and Roger Grosse.
\newblock Picking winning tickets before training by preserving gradient flow.
\newblock In {\em ICLR}, 2020.

\bibitem[\protect\citeauthoryear{Wang \bgroup \em et al.\egroup }{2020b}]{wang2020structured}
Ziheng Wang, Jeremy Wohlwend, and Tao Lei.
\newblock Structured pruning of large language models.
\newblock In {\em EMNLP}, 2020.

\bibitem[\protect\citeauthoryear{Workshop}{2023}]{workshop2023bloom}
BigScience Workshop.
\newblock {BLOOM}: A 176b-parameter open-access multilingual language model.
\newblock {\em arXiv preprint arXiv:2211.05100}, 2023.

\bibitem[\protect\citeauthoryear{Wu \bgroup \em et al.\egroup }{2020}]{wu2020debiased}
Yiquan Wu, Kun Kuang, Yating Zhang, Xiaozhong Liu, Changlong Sun, Jun Xiao, Yueting Zhuang, Luo Si, and Fei Wu.
\newblock De-biased court’s view generation with causality.
\newblock In {\em EMNLP}, pages 763--780, 2020.

\bibitem[\protect\citeauthoryear{Wu \bgroup \em et al.\egroup }{2023}]{wu2023survey}
Likang Wu, Zhi Zheng, Zhaopeng Qiu, et~al.
\newblock A survey on large language models for recommendation.
\newblock {\em arXiv preprint arXiv:2305.19860}, 2023.

\bibitem[\protect\citeauthoryear{Xia \bgroup \em et al.\egroup }{2024}]{xia2024sheared}
Mengzhou Xia, Tianyu Gao, Zhiyuan Zeng, and Danqi Chen.
\newblock {Sheared LLaMA}: Accelerating language model pre-training via structured pruning.
\newblock In {\em ICLR}, 2024.

\bibitem[\protect\citeauthoryear{Xiao \bgroup \em et al.\egroup }{2023}]{xiao2023smoothquant}
Guangxuan Xiao, Ji~Lin, Mickael Seznec, Hao Wu, Julien Demouth, and Song Han.
\newblock {SmoothQuant}: Accurate and efficient post-training quantization for large language models.
\newblock In {\em ICML}, 2023.

\bibitem[\protect\citeauthoryear{Yu \bgroup \em et al.\egroup }{2022}]{yu2022combinatorial}
Xin Yu, Thiago Serra, Srikumar Ramalingam, and Shandian Zhe.
\newblock The combinatorial brain surgeon: Pruning weights that cancel one another in neural networks.
\newblock In {\em ICML}, 2022.

\bibitem[\protect\citeauthoryear{Zellers \bgroup \em et al.\egroup }{2019}]{zellers2019hellaswag}
Rowan Zellers, Ari Holtzman, Yonatan Bisk, Ali Farhadi, and Yejin Choi.
\newblock {Hellaswag}: Can a machine really finish your sentence?
\newblock In {\em ACL}, 2019.

\bibitem[\protect\citeauthoryear{Zhang \bgroup \em et al.\egroup }{2022}]{zhang2022opt}
Susan Zhang, Stephen Roller, Naman Goyal, et~al.
\newblock {OPT}: Open pre-trained transformer language models.
\newblock {\em arXiv preprint arXiv:2205.01068}, 2022.

\bibitem[\protect\citeauthoryear{Zhou \bgroup \em et al.\egroup }{2022}]{zhou2022transpim}
Minxuan Zhou, Weihong Xu, Jaeyoung Kang, and Tajana Rosing.
\newblock {TransPIM}: A memory-based acceleration via software-hardware co-design for transformer.
\newblock In {\em HPCA}, 2022.

\bibitem[\protect\citeauthoryear{Zhu \bgroup \em et al.\egroup }{2015}]{zhu2015aligning}
Yukun Zhu, Ryan Kiros, Rich Zemel, Ruslan Salakhutdinov, Raquel Urtasun, Antonio Torralba, and Sanja Fidler.
\newblock Aligning books and movies: Towards story-like visual explanations by watching movies and reading books.
\newblock In {\em ICCV}, 2015.

\end{thebibliography}
\end{document}